\documentclass[lettersize,journal]{IEEEtran}	
	\usepackage{cite}
	\usepackage{amsmath}
	\usepackage{graphicx}
	\usepackage{textcomp}
	\usepackage{xcolor}	
	\usepackage{float}
	\usepackage{wrapfig}
	\usepackage{algpseudocode}
	\usepackage{multirow}
	\usepackage{amsfonts}
	\usepackage{mathrsfs}	
	\usepackage{amssymb,latexsym}
	\usepackage{commath}
	\usepackage{caption}
	\usepackage{subcaption}	
	\usepackage{booktabs}
	\usepackage{mathtools}	
	\usepackage{todonotes}
	\usepackage{soul}
	\usepackage[ruled,vlined]{algorithm2e}
	\usepackage{multicol}
	\usepackage{amsbsy}
	\usepackage{comment}
	\usepackage[T1]{fontenc}
	\usepackage{amsmath}
	\usepackage[cmintegrals]{newtxmath}
	\usepackage{bm}
	\usepackage{array}
	\usepackage{url}
	\newcommand{\vect}[1]{\boldsymbol{#1}}
	\graphicspath{{Figs/}}
    \usepackage{multicol}
	\usepackage{amsmath}
	\usepackage{graphicx}
	\usepackage{enumerate}
	\usepackage{url}
	\usepackage{longtable}

\begin{document}

\title{Physics-augmented Multi-task Gaussian Process for Modeling Spatiotemporal Dynamics}

\author{Xizhuo Zhang and Bing Yao$^*$
\thanks{
	Xizhuo Zhang and Bing Yao ($^*$corresponding author: byao3@utk.edu) are with the Department of Industrial and Systems Engineering, The University of Tennessee, Knoxville, TN 37996 USA 
	}
}



\maketitle

\begin{abstract}
  Recent advances in sensing and imaging technologies have enabled the collection of high-dimensional spatiotemporal data across complex geometric domains. However, effective modeling of such data remains challenging due to irregular spatial structures, rapid temporal dynamics, and the need to jointly predict multiple interrelated physical variables. This paper presents a physics-augmented multi-task Gaussian Process (P-M-GP) framework tailored for spatiotemporal dynamic systems. Specifically, we develop a geometry-aware, multi-task Gaussian Process (M-GP) model to effectively capture intrinsic spatiotemporal structure and inter-task dependencies.  To further enhance the model fidelity and robustness, we incorporate governing physical laws through a physics-based regularization scheme, thereby constraining predictions to be consistent with governing dynamical principles. We validate the proposed P-M-GP framework on a 3D cardiac electrodynamics modeling task. Numerical experiments demonstrate that our method significantly improves prediction accuracy over existing methods by effectively incorporating domain-specific physical constraints and geometric prior. 
\end{abstract}

\def\abstractname{Note to Practitioners}
\begin{abstract}
    This article proposes a P-M-GP framework designed for predictive modeling of spatiotemporal systems over complex geometric domains. A central feature of the proposed approach is its ability to capture interrelated physical variables while respecting underlying dynamical laws.  This framework can provide more accurate and physically consistent predictions even when available data are limited or irregularly distributed. Although our case study focuses on 3D cardiac electrodynamics, the framework is broadly applicable to other engineering and scientific domains where accurate spatiotemporal prediction is critical, such as environmental monitoring, structural integrity assessment, fluid dynamics, and advanced manufacturing. By incorporating both physics-based regularization and geometry priors, the framework achieves higher predictive fidelity and greater robustness in the presence of sparse or noisy measurements. These capabilities enable reliable decision support in real-world systems where the interplay of geometry and physics is indispensable.
\end{abstract} 

\begin{IEEEkeywords}
  Physics-augmented Modeling, Multi-task Modeling, Spatiotemporal Gaussian Process, Cardiac Electrodynamics.
\end{IEEEkeywords}

\section{Introduction} \label{s:intro}

  Spatiotemporal dynamic processes are fundamental to a wide array of scientific and engineering disciplines \cite{stroud2001dynamic}, governing complex phenomena such as cardiac electrophysiology \cite{yang2023sensing, trayanova2011whole,sim2020epicardial}, thermal-fluid dynamics in additive manufacturing \cite{yao2018multifractal, wang2024sub}, and soil moisture and salinity transport in agricultural farmlands \cite{xie2024effect}. These processes are inherently characterized by interactions among system components evolving across space and time, often unfolding over unstructured 3D geometries with complex spatial-temporal dependencies. Additionally, many real-world spatiotemporal systems are inherently multi-physics, where distinct physical phenomena interact to drive system dynamics. For example,  in cardiac electrodynamics, the propagation of electrical signals is intrinsically coupled with the tissue’s conductivity, the cell's transmembrane potential, and recovery electric current, forming a multi-physics process that spans over the heart geometry \cite{trayanova2011whole}. In additive manufacturing, fabricating 3D structures involves tightly coupled thermal and mechanical processes, all of which directly impact the microstructural evolution and quality of the final part \cite{thompson2015overview}.


Recent advances in sensing and imaging technologies have revolutionized our ability to collect spatiotemporal signals, providing unprecedented visibility into complex dynamic systems \cite{yao2021constrained,wang2021knowledge}. For example, in cardiac electrophysiology, modern multi-lead electrocardiograms (ECGs) and intracardiac catheter mapping techniques enable real-time monitoring of electrical signals both on the torso surface and within cardiac chambers, thereby informing personalized treatment strategies for arrhythmias \cite{yao2024multi,yao2020spatiotemporal}. In AM processes, thermal and optical sensing tools such as infrared cameras and photodiodes are used to track layer-wise temperature fields, melt pool dynamics, and phase transitions, offering insights critical for real-time quality control \cite{yang2019internet}. The massive sensor signals enhance the multifaceted information visibility and can be leveraged to facilitate a better understanding of dynamic systems.


Moreover, many spatiotemporal systems are governed by well-established physical laws, e.g., reaction-diffusion equations, advection-diffusion dynamics, or Navier–Stokes equations, which offer a rich foundation for integrating mechanistic insights into data-driven modeling \cite{karniadakis2021physics}. In cardiac systems, the propagation of electrical potentials is described by nonlinear reaction-diffusion partial differential equations (PDEs) that couple with anatomical and electrophysiological parameters \cite{yang2023sensing,trayanova2011whole}. In AM processes, the transient temperature fields follow the heat conduction equation, with process-induced variability affecting melt pool dynamics and part quality \cite{guo2022deep,ko2023framework}. Deviations in these physical processes often signal system anomalies or failures, such as arrhythmic events in cardiac tissue due to disrupted electrical propagation. Thus, effective system monitoring, anomaly detection, and predictive control critically depend on developing unified modeling frameworks that integrate data-driven learning with physics-based priors, enabling robust and accurate prediction. 

 This paper presents a physics-augmented multi-task Gaussian Process (P-M-GP) framework for predictive modeling of multi-physics spatiotemporal dynamic systems. Our specific contributions are listed as follows:

 (1) Many spatiotemporal systems are defined over complex geometries including asymmetric structures, non-uniform meshes, and curved manifolds \cite{chen2016numerical, yao2016mesh, calandra2016manifold, xie2024kronecker, clayton2011models,10824917}, which pose substantial obstacles for conventional modeling approaches. Additionally, real-world systems often involve multiple interconnected tasks, necessitating predictive methods that can simultaneously model diverse variables \cite{caruana1997multitask}, and further enable downstream decision-making tasks such as anomaly detection. We adapt the geometry-aware spatiotemporal Gaussian Process (G-ST-GP) framework \cite{yao2025geometry} to effectively incorporate geometric information of 3D systems into the spatiotemporal modeling, and extend it with a multi-task Gaussian Process (M-GP) model to capture cross-task dependencies. Additionally, we enhance computational scalability by exploiting the Kronecker product structure of the spatiotemporal kernel, enabling efficient posterior predictions through optimized matrix operations. 
 
 (2) Pure data-driven approaches often neglect the fundamental physical principles governing spatiotemporal systems. This disregard can result in predictions that violate inherent physical constraints, exhibit poor generalizability beyond training distributions, or fail under sparse data conditions where prior knowledge of system dynamics is critical \cite{karpatne2017theory}. We propose an effective physics-augmented M-GP (P-M-GP) strategy to incorporate the physics prior knowledge of the dynamic systems into M-GP for spatiotemporal predictive modeling. This hybrid approach leverages governing physics (often illustrated as PDEs) to constrain the model predictions for improving both physical fidelity and predictive robustness. 

We validate our P-M-GP framework by applying it to spatiotemporal predictive modeling of cardiac electrodynamics in a 3D ventricular geometry. Numerical experiments highlight the superior prediction performance of our P-M-GP model compared with the standard M-GP and PINN. The remainder of this paper is organized as follows: Section~\ref{s:review} presents the review of spatiotemporal predictive modeling. Section~\ref{s:method} introduces the proposed P-M-GP methodological framework. Section~\ref{s:results} demonstrates the effectiveness of our P-M-GP method through numerical experiments in predictive modeling of cardiac electrodynamics. Finally, Section~\ref{s:conclusions} concludes the present investigation.

\section{Research Background} \label{s:review}
    The rapid advancement in sensing and monitoring technologies has led to an abundance of multi-source data streams, creating new challenges in signal integration and prediction across spatiotemporal domains \cite{yan2018real,guo2020hierarchical,liu2022statistical,xie2022physics,yao2016physics}. 
    Gaussian processes (GPs) \cite{williams2006gaussian,liu2020gaussian} offer a flexible, probabilistic framework for spatiotemporal inference and have been deployed across a wide range of domains, including cardiac electrophysiology, robotics, and advanced manufacturing \cite{hu2020gaussian,wang2018spatial,soh2014spatio}.  For example, Senanayake \textit{et al.} developed a GP regression method to model complex space-time dependencies and predict the spread of influenza \cite{Senanayake_Ramos_2016}. Aftab \textit{et al.} built a spatiotemporal GP model to capture critical features in Internal Reference Point data for dynamic,  extended object tracking \cite{8601344}. Zhang and Yao developed a geometry-aware spatiotemporal GP to effectively integrate the temporal correlations and geometric manifold features for dynamic predictive modeling. While conventional single-task GPs have demonstrated success in modeling individual spatiotemporal processes, they are inherently limited when faced with systems characterized by multiple, interdependent variables. In particular, single-task GPs ignore the underlying correlations among related tasks and prevent information sharing across tasks, resulting in limited predictive accuracy and robustness.
    
    Multi‑task extensions (M‑GPs) further boost accuracy by sharing information across correlated tasks \cite{bonilla2007multi}, enabling successful applications in robotics, environmental sensing, additive manufacturing, and biomedicine \cite{yu2005learning,williams2008multi,osborne2008towards,durichen2014multi,shen2023multi}. Building on these foundations, recent works have extended M-GPs to spatiotemporal domains. For example, Akbari and Zhu integrated multi-output spatiotemporal GP with Kalman Filter for dynamically tracking multiple dependent extended targets \cite{9736401}. Hamelijnck \textit{et al.} developed a multi-resolution M-GP to capture varying sampling resolutions and noise levels for air pollution forecasting \cite{NEURIPS2019_0118a063}. Zhang \textit{et al.} developed an M-GP by capturing shared patterns in weather and socioeconomic factors to predict hourly power loads across six cities \cite{zhang2014power}. Gilanifar \textit{et al.} leveraged the M-GP to jointly model dynamic smart-meter readings, weather data, and traffic patterns from two residential areas in Florida, achieving better prediction accuracy \cite{gilanifar2019multitask}. However, most existing spatiotemporal M-GP approaches assume purely data-driven formulations and ignore the underlying physics-based principles, limiting their applicability to scenarios with sparse observations.

    To address the shortcomings in pure data-driven modeling, researchers increasingly embed mechanistic knowledge, typically expressed as algebraic equations, ordinary differential equations (ODEs), or PDEs, into statistical and machine learning approaches \cite{alber2019integrating,karpatne2017theory,karniadakis2021physics,wang2019modeling,wang2021knowledge}.  For example, Physics-Informed Neural Networks (PINNs) have emerged as a popular framework for embedding physics-based PDEs directly into the learning process of deep neural networks \cite{raissi2019physics}. PINNs have been applied successfully to a range of spatiotemporal problems such as cardiac electrodynamics, fluid flow modeling, and materials deformation \cite{xie2022physics,xie2022physics2,cuomo2022scientific,cai2021physics}. However, PINNs are typically constructed using standard feedforward architectures defined on Euclidean domains (e.g., rectangular grids). Extending PINNs to handle spatiotemporal data distributed over irregular, non-Euclidean geometries such as curved manifolds and non-uniform meshes remains challenging \cite{costabal2024delta}. Additionally, PINNs involve the optimization of a very large number of network parameters, often in the order of hundreds of thousands to millions. This high-dimensional parameter space exacerbates training instability, demands substantial computational resources, and makes convergence sensitive to hyperparameter tuning \cite{krishnapriyan2021characterizing}. Moreover, PINNs inherently provide point estimates without calibrated uncertainty quantification, limiting their reliability for decision-making under data sparsity and noise.

    Physics‑informed GPs represent another powerful approach for incorporating physical knowledge into probabilistic modeling \cite{swiler2020survey}. By treating governing differential operators as linear transformations of GP priors, Raissi and Karniadakis introduced the Hidden Physics Model framework, which can solve nonlinear PDEs from limited observations \cite{raissi2018hidden}. Moreover, they also introduced the concept of numerical GPs which embedded time‑stepping operators directly into the covariance, extending to general transient systems \cite{raissi2018numerical}. Recent advances have cast physics‑informed GP regression as a probabilistic analogue of weighted‑residual/Galerkin schemes, yielding rigorous discretisation‑error estimates \cite{pfortner2022physics}. Moreover, the design of problem-specific covariance kernels has been shown to systematically encode boundary conditions directly into the GP prior, ensuring physical consistency of the inferred solutions \cite{dalton2024boundary}. These developments demonstrate the growing maturity of physics‑informed GPs. However, very little has been done to incorporate physics-based knowledge into M-GPs for high-dimensional predictive modeling of spatiotemporal dynamics within complicated 3D geometries.

\section{Research Methodology} \label{s:method}

\begin{figure*}
	\centering
	\includegraphics[width=0.9\linewidth]{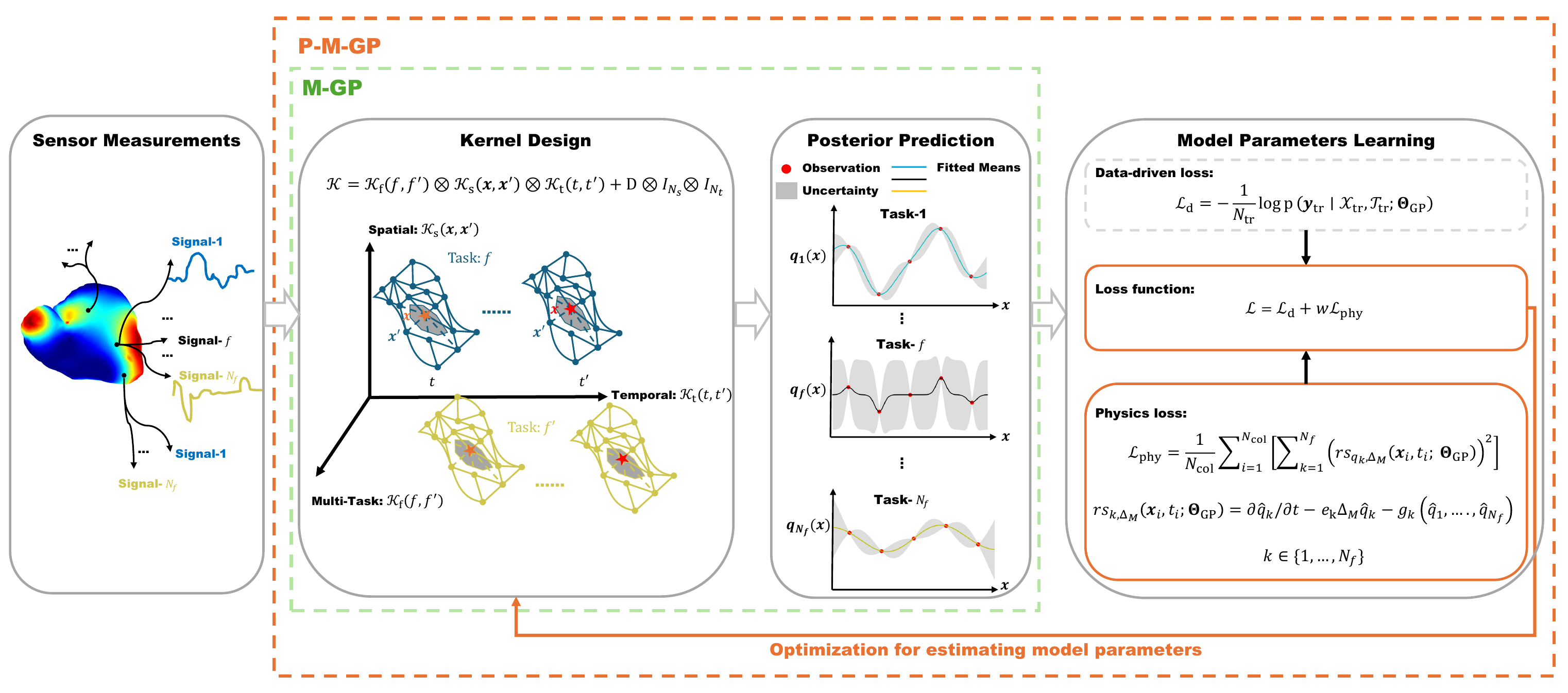}
    \caption{Flowchart of the proposed methodology. The P-M-GP framework is developed to model complex spatiotemporal dynamics by designing effective task-spatial-temporal correlation kernels and integrating physics-based knowledge. The physics knowledge incorporation is achieved by augmenting the physics-based loss, $\mathcal{L}_{\text{phy}}$, into the data-driven loss, $\mathcal{L}_{\text{d}}$, to respect the underlying physics-based principle.}
	\label{fig:flowchart}
\end{figure*}

Fig.~\ref{fig:flowchart} shows the proposed P-M-GP framework. The framework consists of two main components. First, we propose an M-GP to model complex spatiotemporal dynamics through the design of an effective task-spatial-temporal correlation kernel. This kernel captures structural relationships across tasks, spatial locations, and temporal instances by incorporating Laplacian spectral analysis of the 3D geometry and carefully constructed temporal and task covariance functions. Second, a physics-based constraint mechanism is incorporated into the M-GP to enforce physical consistency. This is achieved by augmenting the learning objective with a physics-based loss, yielding the P-M-GP model, to integrate domain knowledge for enhanced predictive accuracy.


\subsection{Spatiotemporal M-GP Modeling}
The spatiotemporal M-GP is designed to enable the integration of multiple tasks within a unified framework for high-dimensional predictive modeling. In our M-GP framework, the variable $\vect{q}(\vect{x},t)\in \mathbb{R}^{N_\text{f}}$ represents the multi-variate system dynamics at a given location $\vect{x}$ and time $t$, with $\vect{y}(\vect{x},t)\in \mathbb{R}^{N_\text{f}}$ denoting the corresponding measurements: 
\begin{equation}
  \begin{aligned}
  \vect{y}(\vect{x},t) &= \vect{q}(\vect{x},t)+\vect{\epsilon}(\vect{x},t) \\
  \vect{q} &\sim \mathcal{GP}(\vect{0},\mathcal{K}_\text{fst})
     \end{aligned}
\end{equation}
where $\vect{\epsilon}(\vect{x},t)$ is a noise term or nugget effect, which follows a multi-variate Gaussian distribution:
      $\vect{\epsilon}(\vect{x},t)\sim \mathcal{N}(\vect{0},D)$,
 where $D$ is a diagonal matrix with elements defined as task-specific noise variance, i.e., $D=\text{diag}(\sigma^2_{1,\epsilon},\dots,\sigma^2_{N_\text{f},\epsilon})$. Additionally, to capture the multi-variate and spatiotemporal interactions, we design the kernel function for $\vect{q}(\vect{x},t)$ as $\mathcal{K_\text{fst}} := \mathcal{K}_{\text{f}} \otimes \mathcal{K}_{\text{s}} \otimes \mathcal{K}_{\text{t}}$, which is a composite function that combines spatial $\mathcal{K}_\text{s}$, temporal $\mathcal{K}_\text{t}$, and task-related $\mathcal{K}_\text{f}$ kernels, each tailored to address specific aspects of correlation inherent in the spatiotemporal multi-physics dynamics. 
 
The training data, represented as $\vect{y}_\text{tr}=[y(f,\vect{x}_i,t_j)]_{f\in F,\vect{x}_i\in\mathcal{X}_\text{tr},t_j\in\mathcal{T}_\text{tr}}\subset \mathbb{R}^{N_\text{f}N_\text{s}N_\text{t}}$, comprises observations collected at specified spatial locations $\mathcal{X}_\text{tr}=\{\vect{x}_1,\dots,\vect{x}_{N_\text{s}}\}$ and time points $\mathcal{T}_\text{tr}=\{t_1,\dots,t_{N_\text{t}}\}$ across different tasks $F=\{f_1, \dots, f_{N_\text{f}}\}$. Within the framework of M-GP modeling, the marginal distribution of the training data is given by:
\begin{equation}
  \vect{y}_\text{tr}\sim\mathcal{N}(\vect{0},\Sigma_\text{tr})
  \label{Eq: marginal}
\end{equation}
with $\Sigma_\text{tr}=\mathcal{K}_\text{f}(F,F)
\otimes
\mathcal{K}_\text{s}(\mathcal{X}_\text{tr},\mathcal{X}_\text{tr})
\otimes
\mathcal{K}_\text{t}(\mathcal{T}_\text{tr},\mathcal{T}_\text{tr})
 + D \otimes I_{N_\text{s}} \otimes I_{N_\text{t}}$ denoting the covariance of $\vect{y}_\text{tr}$, where $I_{N_\text{s}}$ ($I_{N_\text{t}}$) is an $N_\text{s}\times N_\text{s}$ ($N_\text{t}\times N_\text{t}$) identity matrix. Furthermore, given collected observations $[\vect{y}_\text{tr};\mathcal{X}_\text{tr},\mathcal{T}_\text{tr}]$, the predictive distribution of the dynamics for the $f^*$-th task at an arbitrary spatiotemporal coordinate $(\vect{x}^*,t^*)$ is:
\begin{equation}
  \begin{aligned}
    &q(f^*, \vect{x}^*, t^*) \mid \vect{y}_\text{tr}; \mathcal{X}_\text{tr}, \mathcal{T}_\text{tr}
  \sim \mathcal{N} \big( \mu^*, \Sigma^* \big), \label{eq:trivial-GP}  \\
  &\mu^* = \mathcal{K}_\text{fst}^* \Sigma_\text{tr}^{-1} \vect{y}_\text{tr},  \\
  &\Sigma^* = \mathcal{K}_\text{fst}^{**} - \mathcal{K}_\text{fst}^* \Sigma_\text{tr}^{-1} (\mathcal{K}_\text{fst}^*)^\top
  \end{aligned}
\end{equation}
where $ \mathcal{K}_\text{fst}^* =\mathcal{K}_\text{fst}\left((f^*, \vect{x}^*,t^*);(F,\mathcal{X}_\text{tr},\mathcal{T}_\text{tr})\right)$ and $ \mathcal{K}_\text{fst}^{**} =\mathcal{K}_\text{fst}\left((f^*,\vect{x}^*,t^*);(f^*,\vect{x}^*,t^*)\right)$.  As such, the predictive performance fundamentally depends on the construction of effective kernels of $\mathcal{K}_\text{s}$, $\mathcal{K}_\text{t}$ and $\mathcal{K}_\text{f}$, which will be detailed as follows. 


 \textbf{Spatial Kernel Construction}: To effectively capture the geometric features of the complex system, we construct the spatial kernel based on the Laplacian operator of the system geometry, following the approach established in \cite{yao2025geometry}. The Laplacian operator $\Delta_M$ defined on the discretized 3D surface $M$, which comprises $N$ vertices, encodes essential geometric properties such as vertex connectivity, edge orientation, and local curvature. Specifically, for a given vertex $i$, the discrete Laplacian is defined as \cite{sorkine2005laplacian}:
    \begin{equation}
        (\Delta_M \vect{q})_i=\frac{1}{2|\Omega_i|}\sum_{j \in N(i)}(\cot \alpha_{ij} + \cot \beta_{ij})(\vect{q}_j-\vect{q}_i)
        \label{eq:angle}
    \end{equation}
where $|\Omega_i|$ is the Voronoi cell area around vertex $i$, $N(i)$ contains the neighboring vertices connected to $i$, $\alpha_{ij}$ and $\beta_{ij}$ are the angles opposite to the edge between vertices $i$ and $j$. Using the eigenfunctions ${\phi_j(\cdot)}$ and the corresponding eigenvalues ${\lambda_j}$ of $\Delta_M$, we define our spatial kernel as:
    \begin{equation}
        \mathcal{K}_{\text{s}}(\vect{x}, \vect{x}') = \sum_{j \in \mathbb{N}^*} \sigma_{m} S\left( \sqrt{\lambda_j} \right) \, \phi_j(\vect{x}) \, \phi_j(\vect{x}')  
        \label{eq: def-kernel-spa}
    \end{equation}
where $\sigma_m$ is a scaling coefficient and $S(\sqrt{\lambda})$ is a non-increasing spectral density that modulates the contribution of each eigenmode, which is selected as the spectral density of the Matérn covariance structure in the frequency domain:
\begin{equation}
  \begin{aligned}
  S(\sqrt{\lambda}) 
  &:= \frac{2^d \pi^{d/2} \Gamma(\nu + d/2) (2 \nu)^{\nu}}%
           {\Gamma(\nu) l_{\text{s}}^{2 \nu}}\\
  &\quad \times \left( \frac{2 \nu}{l_{\text{s}}^2} + 4 \pi^2 \lambda \right)^{-(\nu + d/2)}
  \label{eq:spectral-density}
\end{aligned}
\end{equation}
where $d=2$ corresponds to the intrinsic dimensionality of the surface manifold, $\nu=3/2$ is the smoothness parameter, $l_\text{s}$ is the spatial length-scale, and $\Gamma(\cdot)$ denotes the Gamma function.

 \textbf{Temporal Kernel Construction}: 
 Similar to the prior work \cite{yao2025geometry}, we adopt the Matérn kernel with the smoothness $\nu=3/2$ to model the temporal evolution of the dynamics: 
\begin{equation}
  \begin{aligned}
      \mathcal{K}_{\text{t}}(t, t') 
      = \sigma_a \left( 1 + \frac{\sqrt{3}\|t - t'\|}{l_{\text{t}}} \right) 
      \exp\left( - \frac{\sqrt{3}\|t - t'\|}{l_{\text{t}}} \right)
      \label{eq:def-kernel-tem}
    \end{aligned}
\end{equation}    
    Here, $l_\text{t}$ is a length scale controlling the temporal range of correlation, while $\sigma_a$ acts as a scaling coefficient adjusting the overall magnitude of temporal variability. This structure ensures that closer time points exhibit stronger correlations, while allowing for smooth decay over longer intervals.

\textbf{Task-Correlation Kernel Construction:} We use a free-form kernel to capture the correlations between multiple tasks. Specifically, we define the task-correlation kernel as:
\begin{equation}
  \begin{aligned}
      \mathcal{K}_{\text{f}} = L L^\top, ~~~
      L &= \begin{bmatrix} 
      \beta_{11} & 0 & \cdots & 0 \\
      \beta_{21} & \beta_{22} & 0 & 0 \\
      \vdots & \vdots & \ddots & \vdots \\
      \beta_{N_\text{f}1} & \beta_{N_\text{f}2} & \cdots & \beta_{N_\text{f}N_\text{f}}
      \end{bmatrix}
      \label{eq:Def-ker}
  \end{aligned}
\end{equation}
where \( L \) is a lower-triangular matrix with learnable parameters \(\{\beta_{ij}\}\)'s. This decomposition guarantees that $\mathcal{K}_{\text{f}}$ is symmetric and positive semi-definite, thus satisfying the requirements of a valid covariance kernel. The free-form structure of \(L\) allows the model to flexibly learn arbitrary task relationships from data without imposing restrictive parametric assumptions. 

\subsection{Posterior Prediction} \label{s:posterior}
 Given the kernel design and M-GP parameters, $\Theta_\text{GP}=\{\sigma_m,l_\text{s},\sigma_a,l_\text{t},\beta_{11},\dots,\beta_{N_\text{f}N_\text{f}}, \sigma_{1,\epsilon}^2, \dots, \sigma_{N_\text{f},\epsilon}^2\}$, the predicted mean of the dynamics is derived as:
\begin{equation}
  \begin{aligned}
      q(f^*,\vect{x}^*, t^*) |_{\vect{y}_{\text{tr}}, \mathcal{X}_{\text{tr}}, \mathcal{T}_{\text{tr}}; \Theta_\text{GP}} 
      = \left( \mathcal{K}_\text{f}^* \otimes \mathcal{K}_\text{s}^* \otimes \mathcal{K}_\text{t}^* \right) \Sigma_{\text{tr}}^{-1} \vect{y}_{\text{tr}}
  \end{aligned}
\end{equation}
where $\mathcal{K}^*$ is the kernel matrix computed between test and training points. $\Sigma_{\text{tr}}^{-1}$ denotes the inverse covariance matrix constructed from the training data, and $\vect{y}_{\text{tr}}$ contains the observed values in the training dataset, stacked as a vector.
According to the kernel design, $\Sigma_{\text{tr}}$ can be written as $\Sigma_{\text{tr}} = \mathcal{K}_{\text{f}} \otimes \Sigma_{\text{tr},\text{s}} \otimes \Sigma_{\text{tr},\text{t}} + D \otimes I_{N_\text{s}} \otimes I_{N_\text{t}}$, where $\Sigma_{\text{tr},\text{s}} :=\mathcal{K}_\text{s}(\mathcal{X}_\text{tr},\mathcal{X}_\text{tr})$, $\Sigma_{\text{tr},\text{t}}:=\mathcal{K}_\text{t}(\mathcal{T}_\text{tr},\mathcal{T}_\text{tr})$. 
In a special case of two tasks (i.e., $F=\{1,2\}$): the task correlation kernel is a $2\times 2$ matrix, i.e., $\mathcal{K}_{\text{f}}(F,F) = LL^\top:= \begin{bmatrix} \beta_{uu} & \beta_{uv} \\ \beta_{uv} & \beta_{vv} \end{bmatrix}$, where we use $u$ and $v$ to denote task 1 and 2, respectively; the corresponding noise matrix is resulted as $D=\text{diag}(\sigma_{u,\epsilon}^2,\sigma_{v,\epsilon}^2)$. To efficiently compute the posterior distribution in our framework, we need to address the computational challenges posed by the large-scale covariance matrices $\Sigma_{\text{tr}}$ involved. Specifically, for a multi-task prediction scenario with $N_\text{f}$ tasks, $N_\text{s}$ spatial points, and $N_\text{t}$ temporal points, $\Sigma_{\text{tr}}$ is with the dimension of $(N_\text{f}N_\text{s}N_\text{t}) \times (N_\text{f}N_\text{s}N_\text{t})$, which can be prohibitively large for direct inversion and determinant computation.

To overcome the computational barrier, we leverage the Kronecker product structure and employ Singular Value Decomposition (SVD) to facilitate efficient matrix operations. By decomposing the spatial covariance matrix $\Sigma_{\text{tr},\text{s}} := U_\text{s} \Lambda_\text{s} U_\text{s}^{\top}$ and temporal covariance matrix $\Sigma_{\text{tr},\text{t}} := U_\text{t} \Lambda_\text{t} U_\text{t}^{\top}$ into their respective orthogonal components via SVD (where both $U_\text{s}$ and $U_\text{t}$ are orthogonal matrices with $U_\text{s}U_\text{s}^\top=I$ and $U_\text{t}U_\text{t}^\top=I$), we can restructure the full covariance matrix $\Sigma_{\text{tr}}$ in a form that allows for more tractable computations:
\begin{equation}
  \begin{aligned}
      \Sigma_{\text{tr}} 
      &= \mathcal{K}_{\text{f}} \otimes \Sigma_{\text{tr},\text{s}} \otimes \Sigma_{\text{tr},\text{t}} 
      + D \otimes I_{N_\text{s}} \otimes I_{N_\text{t}} \\
      &= 
      \begin{bmatrix}
      \Sigma_{11} & \Sigma_{12} \\
      \Sigma_{21} & \Sigma_{22}
      \end{bmatrix}
      := U \Lambda U^\top
      \label{eq:Sigma-decomp}
      \end{aligned}
    \end{equation}
where the intermediate variables are defined as:
\begin{equation}
  \begin{aligned}
  \Sigma_{11} &:= (U_\text{s} \otimes U_\text{t}) 
  \left[ \beta_{uu} (\Lambda_\text{s} \otimes \Lambda_\text{t}) \right] 
  (U_\text{s} \otimes U_\text{t})^\top   \\
  &\quad + \sigma_{u,\epsilon}^2 
  (U_\text{s} \otimes U_\text{t}) (I_\text{s} \otimes I_\text{t}) (U_\text{s} \otimes U_\text{t})^\top   \\[0.5em]
  \Sigma_{21} &= \Sigma_{12} := (U_\text{s} \otimes U_\text{t}) 
  \left[ \beta_{uv} (\Lambda_\text{s} \otimes \Lambda_\text{t}) \right] 
  (U_\text{s} \otimes U_\text{t})^\top   \\[0.5em]
  \Sigma_{22} &:= (U_\text{s} \otimes U_\text{t}) 
  \left[ \beta_{vv} (\Lambda_\text{s} \otimes \Lambda_\text{t}) \right] 
  (U_\text{s} \otimes U_\text{t})^\top   \\
  &\quad + \sigma_{v,\epsilon}^2 
  (U_\text{s} \otimes U_\text{t}) (I_\text{s} \otimes I_\text{t}) (U_\text{s} \otimes U_\text{t})^\top   \\
  U &:= \mathrm{diag}\left(U_\text{s} \otimes U_\text{t},\; U_\text{s} \otimes U_\text{t} \right)   \\
  \Lambda &:= \begin{bmatrix}
    \Lambda_{11} & \Lambda_{12} \\
    \Lambda_{21} & \Lambda_{22}
  \end{bmatrix}
    \end{aligned}
\end{equation}
where:
\begin{equation}
  \begin{aligned}
\Lambda_{11} &:= \beta_{uu} (\Lambda_\text{s} \otimes \Lambda_\text{t}) 
+ \sigma_{u,\epsilon}^2 (I_\text{s} \otimes I_\text{t}) \\
\Lambda_{12} &= \Lambda_{21} := \beta_{uv} (\Lambda_\text{s} \otimes \Lambda_\text{t}) \\
\Lambda_{22} &:= \beta_{vv} (\Lambda_\text{s} \otimes \Lambda_\text{t}) 
+ \sigma_{v,\epsilon}^2 (I_\text{s} \otimes I_\text{t})
\end{aligned}
\end{equation}
Note that $\Lambda$ can be regarded as a $2 \times 2$ block matrix, defined as $\Lambda = \begin{bmatrix} \Lambda_{11} & \Lambda_{12} \\\Lambda_{12} & \Lambda_{22} \end{bmatrix}$ with $\Lambda_{11}$, $\Lambda_{12}$ and $\Lambda_{22}$ being diagonal matrices, and $U$ is an orthogonal matrix with $UU^\top=I$.

By leveraging our decomposition approach, we can express $\Sigma_{\text{tr}}^{-1}$ in terms of smaller, more manageable matrices, thereby facilitating efficient computation of the posterior mean and variance:
    \begin{equation}
  \begin{aligned}
      (\Sigma_{\text{tr}})^{-1} 
      &= (U \Lambda U^\top)^{-1} 
      = (U^\top)^{-1} \Lambda^{-1} U^{-1}   \\
      &= U \Lambda^{-1} U^\top
      \label{eq:Sigma-tr-inv}
          \end{aligned}
\end{equation}   
where:
 \begin{equation}
  \begin{aligned}
          \Lambda^{-1} :&= \begin{bmatrix}
            \tilde{\Lambda}_{11} & \tilde{\Lambda}_{12} \\
            \tilde{\Lambda}_{12} & \tilde{\Lambda}_{22}
        \end{bmatrix}   \\
          \tilde{\Lambda}_{11} : &= \Lambda_{11}^{-1} + \Lambda_{11}^{-1} \Lambda_{12} S^{-1} \Lambda_{21} \Lambda_{11}^{-1},  \\
          \tilde{\Lambda}_{12} : &= -\Lambda_{11}^{-1} \Lambda_{12} S^{-1},  ~~~
          \tilde{\Lambda}_{22} : = S^{-1},  \\
          S : &= \Lambda_{22} - \Lambda_{12} \Lambda_{11}^{-1} \Lambda_{12}.
          \label{eq:Sigma-tr-inv-defs}
     \end{aligned}
\end{equation}
Here, we use block matrix inversion techniques to decompose $\Lambda^{-1}$ into components that can be computed with operations of diagonal matrices, further enhancing computational efficiency.   

Based on the computation of $\Sigma_{\text{tr}}^{-1}$ (Eqs.~\eqref{eq:Sigma-tr-inv}-\eqref{eq:Sigma-tr-inv-defs}), we can now derive efficient expressions for the posterior mean through a series of structured operations as follows:
\begin{equation}
  \begin{aligned}
      &\vect{q}(\vect{x}^*, t^*) |_{\vect{y}_{\text{tr}}, \mathcal{X}_{\text{tr}}, \mathcal{T}_{\text{tr}}; \Theta_\text{GP}}  \\
      &= \left( \mathcal{K}_\text{f}^* \otimes \mathcal{K}_\text{s}^* \otimes \mathcal{K}_\text{t}^* \right) U \Lambda^{-1} U^\top \vect{y}_{\text{tr}}  \\
      &= \left( \mathcal{K}_\text{f}^* \otimes \mathcal{K}_\text{s}^* \otimes \mathcal{K}_\text{t}^* \right) 
      U \tilde{\Lambda} U^\top
      \begin{bmatrix} \vect{y}_u \\ \vect{y}_v \end{bmatrix}  \\
      &= \left( \mathcal{K}_\text{f}^* \otimes \mathcal{K}_\text{s}^* \otimes \mathcal{K}_\text{t}^* \right) 
      (I_{N_\text{f}} \otimes U_\text{s} \otimes U_\text{t})
      \tilde{\Lambda} \begin{bmatrix} \vect{v}_u \\ \vect{v}_v \end{bmatrix}  \\
      &= \left[ (\mathcal{K}_\text{f}^* I_{N_\text{f}}) \otimes (\mathcal{K}_\text{s}^* U_\text{s}) \otimes (\mathcal{K}_\text{t}^* U_\text{t}) \right]
      \tilde{\Lambda}
      \begin{bmatrix} \vect{v}_u \\ \vect{v}_v \end{bmatrix}  \\
      &= \left[ (\mathcal{K}_\text{f}^* I_{N_\text{f}}) \otimes (\mathcal{K}_\text{s}^* U_\text{s}) \otimes (\mathcal{K}_\text{t}^* U_\text{t}) \right] \tilde{\vect{y}}  \\
      &= \operatorname{vec} \left\{ \operatorname{vec}^{-1} (\tilde{\vect{y}}) \times _1 (\mathcal{K}_\text{t}^* U_\text{t}) \times_2 (\mathcal{K}_\text{s}^* U_\text{s}) \times_3(\mathcal{K}_\text{f}^*) \right\}
      \label{eq:posterior-distribution}
      \end{aligned}
\end{equation}
where:
  \begin{equation}
  \begin{aligned}
  U &= \begin{bmatrix} U_\text{s} \otimes U_\text{t} & 0 \\ 0 & U_\text{s} \otimes U_\text{t} \end{bmatrix} \\
  \tilde{\Lambda} &:= \Lambda^{-1} = 
  \begin{bmatrix}
  \tilde{\Lambda}_{11} & \tilde{\Lambda}_{12} \\
  \tilde{\Lambda}_{21} & \tilde{\Lambda}_{22}
  \end{bmatrix}\\
  \vect{v}_u &:= (U_\text{s} \otimes U_\text{t})^\top \vect{y}_u, \\
  \vect{v}_v &:= (U_\text{s} \otimes U_\text{t})^\top \vect{y}_v, \\
  \tilde{\vect{y}} &:= 
  \begin{bmatrix}
  \tilde{\Lambda}_{11} \vect{v}_u + \tilde{\Lambda}_{12} \vect{v}_v \\
  \tilde{\Lambda}_{21} \vect{v}_u + \tilde{\Lambda}_{22} \vect{v}_v
  \end{bmatrix}
     \end{aligned}
\end{equation}
$\operatorname{vec}(\cdot)$ is the vectorization operator that stacks matrix columns into a single column vector, $\text{vec}^{-1}(\cdot)$ is the inverse of $\text{vec}(\cdot)$ operation and $\text{vec}^{-1}(\tilde{\vect{y}})\in\mathbb{R}^{N_\text{t}\times N_\text{s}\times N_\text{f}}$, and $\times_n$ is the mode-$n$ tensor multiplication.
By exploiting tensor algebra and matrix decomposition techniques, the proposed computational strategies circumvent the direct inversion of the full covariance matrix \( \Sigma_{\text{tr}} \in \mathbb{R}^{(N_\text{f} N_\text{s} N_\text{t}) \times (N_\text{f} N_\text{s} N_\text{t})} \), which is computationally prohibitive for large-scale problems. Instead, the computation of the predictive mean is reduced to a sequence of smaller matrix operations involving matrices of dimensions \( N_\text{f} \times N_\text{f} \), \( N_\text{s} \times N_\text{s} \), and \( N_\text{t} \times N_\text{t} \). This approach enables scalable and efficient high-resolution spatiotemporal modeling across multiple tasks, making it feasible to perform inference on systems where conventional methods would be computationally intractable.



\subsection{Physics-augmented M-GP} \label{s:methods}

\subsubsection{Physics-based Spatiotemporal Reaction-diffusion Model}


The present study focuses on spatiotemporal predictive modeling of a two-variable reaction-diffusion model defined on complex geometries. The evolution of the system dynamics on the 3D manifold $M$ is typically governed by the following set of PDEs:
  \begin{equation}
   \begin{aligned}
      &\frac{\partial u}{\partial t} = e_1 \Delta u + g_1(u,v), \\
      &\frac{\partial v}{\partial t} = e_2 \Delta v + g_2(u,v),\\
      &\vect{n}\cdot \nabla u|_M =0 \\
      &\vect{n}\cdot \nabla v|_M =0
   \end{aligned}
  \end{equation}
  where $u$ and $v$ are the state variables representing the spatiotemporal concentrations of two interacting dynamics, $e_1$ and $e_2$ are the respective diffusion coefficients, $g_1(\cdot)$ and $g_2(\cdot)$ denote the nonlinear reactions kinetics. The boundary conditions (last two equations) enforce zero-flux/gradient (Neumann) constraints on the surface $M$, ensuring that the dynamics are confined to the manifold and there is no flow of material across its boundary.

\subsubsection{Physics-Augmented Learning}

To integrate prior physical knowledge into the GP modeling framework, we propose a physics-augmented parameter inference formulation that jointly leverages observational data and physical constraints. Specifically, the GP hyperparameter estimation is cast as the following regularized optimization problem:
\begin{equation}
  \begin{aligned}
    \Theta_\text{GP}^* =& \arg \min_{\Theta_\text{GP}} \left[\mathcal{L}_\text{d} + w \mathcal{L}_{\text{phy}}\right] \\
    \mathcal{L}_\text{d}:=&
  -\frac{1}{N_{\text{tr}}} \log p(\vect{y}_{\text{tr}} |\mathcal{X}_{\text{tr}}, \mathcal{T}_{\text{tr}}; \Theta_\text{GP})
  \end{aligned}
\end{equation}
where $\mathcal{L}_\text{d}$ represents the data-driven negative log-likelihood that quantifies model fit to observed data, while $\mathcal{L}_{\text{phy}}$ encodes physics-based constraints, derived from governing equations (detailed below), which is introduced as a soft penalty modulated by a weight parameter $w$.

 \textbf{Data-driven M-GP Likelihood}:
  Given the training dataset $[\vect{y}_\text{tr};\mathcal{X}_\text{tr},\mathcal{T}_\text{tr}]$ and the marginal distribution in Eq. (\ref{Eq: marginal}), the negative log-likelihood is expressed as:
  \begin{equation}
  \begin{aligned}
    &-\log p(\vect{y}_\text{tr} \mid \mathcal{X}_\text{tr}, \mathcal{T}_\text{tr}; \Theta_\text{GP})  \\
    &= \frac{1}{2} \vect{y}_\text{tr}^\top \Sigma_\text{tr}^{-1} \vect{y}_\text{tr} 
    + \frac{1}{2} \log \left| \Sigma_\text{tr} \right|  + \frac{N_\text{f} N_\text{s} N_\text{t}}{2} \log (2\pi)    \\
    &=: \frac{1}{2}(f_1 + f_2 + f_3)
    \label{eq:NLL}
        \end{aligned}
\end{equation}   
where $N_\text{f}N_\text{s}N_\text{t}$ is the total number of observations across all tasks ($N_\text{f}$), spatial locations ($N_\text{s}$), and time points ($N_\text{t}$). The likelihood function consists of three components: a data fit term $f_1$ that measures how well the model explains observed data, a complexity penalty $f_2$ that prevents overfitting by controlling model flexibility, and a normalization constant $f_3$. Computing terms $f_1$ and $f_2$ presents significant computational challenges due to the large-scale matrix operations involved. To address these challenges, we employ efficient matrix decomposition strategies that exploit the structural properties of $\Sigma_\text{tr}$ to significantly reduce both computational complexity and memory requirements. Specifically, for $f_1$, we leverage the decomposition of $\Sigma_{\text{tr}}^{-1}$ (from Eq.~\eqref{eq:Sigma-tr-inv} and Eq.~\eqref{eq:Sigma-tr-inv-defs}) to obtain:
  \begin{equation}
  \begin{aligned}
    &f_1 := \vect{y}_{\text{tr}}^\top \Sigma_{\text{tr}}^{-1} \vect{y}_{\text{tr}}
    = \vect{y}_{\text{tr}}^\top U \Lambda^{-1} U^\top \vect{y}_{\text{tr}}   \\
    &= 
    \begin{bmatrix} \vect{y}_u^\top & \vect{y}_v^\top \end{bmatrix}
    \left( I_2 \otimes U_\text{s} \otimes U_\text{t} \right)
    \tilde{\Lambda}
    \left( I_2 \otimes U_\text{s} \otimes U_\text{t} \right)^\top
    \begin{bmatrix} \vect{y}_u \\ \vect{y}_v \end{bmatrix}
    \label{eq:f1-quadratic}
        \end{aligned}
\end{equation}
    
By defining transformed variables $\vect{v}_u := (U_\text{s} \otimes U_\text{t})^\top \vect{y}_u = \operatorname{vec} \left( U_\text{t}^\top \operatorname{vec}^{-1} (\vect{y}_u) U_\text{s} \right)$ and $\vect{v}_v := (U_\text{s} \otimes U_\text{t})^\top \vect{y}_v = \operatorname{vec} \left( U_\text{t}^\top \operatorname{vec}^{-1} (\vect{y}_v) U_\text{s} \right)$, we can simplify this expression to:
  \begin{equation}
    \begin{aligned}
        f_1 &= \begin{bmatrix} \vect{v}_u^\top, \vect{v}_v^\top \end{bmatrix}
        \begin{bmatrix} \tilde{\Lambda}_{11} & \tilde{\Lambda}_{12} \\ \tilde{\Lambda}_{12} & \tilde{\Lambda}_{22} \end{bmatrix}
        \begin{bmatrix} \vect{v}_u \\ \vect{v}_v \end{bmatrix} \\
        &= \vect{v}_u^\top \tilde{\Lambda}_{11} \vect{v}_u + \vect{v}_u^\top \tilde{\Lambda}_{12} \vect{v}_v + \vect{v}_v^\top \tilde{\Lambda}_{12} \vect{v}_u + \vect{v}_v^\top \tilde{\Lambda}_{22} \vect{v}_v
    \end{aligned}
  \end{equation}   

Similarly, for $f_2$, we exploit the properties of orthogonal matrices and block-diagonal structures:
    \begin{equation}
  \begin{aligned}
      &f_2 := \log |\Sigma_{\text{tr}}| 
      = \log |U \Lambda U^\top| 
      = \log |U \Lambda U^{-1}|   \\
      &= \log (|U|\, |\Lambda|\, |U^{-1}|) 
      = \log (|U U^{-1}|\, |\Lambda|) 
      = \log |\Lambda|   \\
      &= \log 
      \begin{vmatrix} 
      \Lambda_{11} & \Lambda_{12} \\ 
      \Lambda_{12} & \Lambda_{22} 
      \end{vmatrix} 
      = \log \left( |\Lambda_{11}| \cdot 
      |\Lambda_{22} - \Lambda_{12} \Lambda_{11}^{-1} \Lambda_{12}| \right)   \\
      &= \log |\Lambda_{11}| 
      + \log |\Lambda_{22} - \Lambda_{12} \Lambda_{11}^{-1} \Lambda_{12}|   \\
      &= \sum_i \log (\Lambda_{11})_{ii} 
      + \sum_i \log \left( \left[\Lambda_{22} - \Lambda_{12} 
      \Lambda_{11}^{-1} \Lambda_{12} \right]_{ii} \right)
      \label{eq:f2-logdet}
          \end{aligned}
\end{equation}      
   It is worth noting that the block matrices, $\Lambda_{11}$, $\Lambda_{12}$, and $\Lambda_{22}$ are all diagonal, which allows for a more efficient computation of the determinant compared to directly evaluating the determinant of the original $(N_\text{f}N_\text{s}N_\text{t})\times(N_\text{f}N_\text{s}N_\text{t})$ matrix $\Sigma_{\text{tr}}$.
    

  \textbf{Physics-based Loss}: To incorporate physics-based constraints into our model, we define residuals that quantify the deviation between the M-GP predictions and the governing reaction-diffusion PDEs:
    \begin{equation}
  \begin{aligned}
      \gamma_{S_u} (\vect{x},t \, ; \, \Theta_\text{GP}) :&= \frac{\partial \hat{u}}{\partial t} -e_1 \Delta \hat{u}-g_1(\hat{u},\hat{v}), \\
      \gamma_{S_v} (\vect{x},t \, ; \, \Theta_\text{GP}) :&= \frac{\partial \hat{v}}{\partial t}-e_2 \Delta \hat{v}-g_2(\hat{u},\hat{v})\\
      \gamma_{u,bc} (\vect{x},t \, ; \, \Theta_\text{GP}) :&= \vect{n}\cdot \nabla \hat{u}|_M \\
      \gamma_{v,bc} (\vect{x},t \, ; \, \Theta_\text{GP}) :&= \vect{n}\cdot \nabla \hat{v}|_M
  \end{aligned}
\end{equation}
These residuals measure how well the M-GP predictions $\hat{u}$ and $\hat{v}$ satisfy the PDEs at any given spatial location and time point. 
Note that to evaluate the physics-based loss, we need to compute the derivatives that appear in the residuals. Specifically, the time derivatives can be computed as:
\begin{equation}
  \begin{aligned}
  \left[
  \frac{\partial \hat{u}}{\partial t},\;
  \frac{\partial \hat{v}}{\partial t}
  \right]^{\operatorname{T}} 
  = \frac{\partial \hat{\vect{q}}}{\partial t} 
  &= \operatorname{vec} \Big\{ 
  \operatorname{vec}^{-1} (\tilde{\vect{y}}) 
  \times_1 \left( \frac{\partial \mathcal{K}_\text{t}^*}{\partial t} U_\text{t} \right)   \\
  &\quad \times_2 (\mathcal{K}_\text{s}^* U_\text{s}) 
  \times_3 (\mathcal{K}_\text{f}^* I_{N_\text{f}}) 
  \Big\}
  \label{eq:time-derivative}
      \end{aligned}
\end{equation}
  
For the Matérn temporal kernel with smoothness parameter $\nu=3/2$, we have:
\begin{equation}
  \begin{aligned}
  &\left[ \frac{\partial \mathcal{K}_t^*}{\partial t} \right]_{i=1 \sim N_{\text{col}},\, j=1 \sim N_{\text{tr}}}   \\
  &= -\sigma_a a^2\, \exp(-a m)\, m\cdot \operatorname{sign}(t_i - t_j)
  \label{eq:dK_dt}
      \end{aligned}
\end{equation} 
where $N_{\text{col}}$ is the number of spatiotemporal collocation points used to enforce the physical constraints, $m = \|t_i - t_j\|$, and $a = \frac{\sqrt{3}}{l_t}$ (see the detailed derivation in the Appendix).

   The spatial derivative in the reaction-diffusion system is represented by the Laplacian operator $\Delta$, which by default operates in the ambient Euclidean space $\mathbb{R}^3$. However, because the dynamics are restricted to a 3D surface $M \subset \mathbb{R}^3$, it is essential to consider the intrinsic geometry of $M$ to derive the spatial derivative. To reconcile this, we leverage the Neumann boundary condition imposed on the state variable $u$ (or $v$) to prove that the Euclidean Laplacian $\Delta u$ restricted to the surface $M$ is equivalent to the surface Laplacian  $\Delta_M u$. Specifically, using the decomposition of the Euclidean gradient and divergence into tangential and normal components, we obtain: 
   \begin{equation}
        \Delta u|_M=\nabla\cdot \nabla_M u+ \nabla\cdot ((\vect{n}\cdot \nabla u|_M)\vect{n})
    \end{equation}
    Because $\nabla_M u$ is tangential to surface $M$ and $\vect{n}\cdot \nabla u|_M=0$ (i.e., the Neumann boundary condition), we conclude that $\Delta u|_M=\nabla_M\cdot\nabla_Mu=\Delta_M u$. As such, the physics-based loss function is then defined as the mean squared residual constructed by the temporal derivative and spatial derivative encoded by the surface Laplacian $\Delta_M$ across a set of collocation points:
\begin{equation}
  \begin{aligned}
  \mathcal{L}_{\text{phy}} 
  &= \frac{1}{N_{\text{col}}} \sum_{i=1}^{N_{\text{col}}} \bigg[ 
  \left( \gamma_{S_u,\Delta_M} (\vect{x}_i, t_i; \Theta_\text{GP}) \right)^2   \\
  &\qquad\qquad\quad + \left( \gamma_{S_v,\Delta_M} (\vect{x}_i, t_i; \Theta_\text{GP}) \right)^2 
  \bigg]
  \label{eq:phy-loss}
      \end{aligned}
\end{equation}
where $\gamma_{S_u,\Delta_M} (\vect{x},t \, ; \, \Theta_\text{GP}) := \frac{\partial \hat{u}}{\partial t} -e_1 \Delta_M \hat{u}-g_1(\hat{u},\hat{v})$ and $\gamma_{S_v,\Delta_M} (\vect{x},t \, ; \, \Theta_\text{GP}) := \frac{\partial \hat{v}}{\partial t} -e_2 \Delta_M \hat{v}-g_2(\hat{u},\hat{v})$.

 \textbf{Efficient Cross-Validation for Model Parameter Selection}: To select optimal model parameters under different initial guesses and evaluate prediction performance, we employ the leave-one-location-out cross-validation. The cross-validation error is defined as:
\begin{equation}
  \begin{aligned}
      \hat{\tau}_{n,\text{cv}}^2 = \frac{1}{n} \sum_{f=1}^{N_\text{f}} \sum_{i=1}^{N_\text{s}} \sum_{j=1}^{N_\text{t}} \left( y(f, \vect{x}_i,t_j) - \hat{q}_{-\vect{x}_i}(f, \vect{x}_i,t_j) \right)^2 \label{eq:tau_square}
  \end{aligned}
\end{equation}
where $\hat{q}_{-\vect{x}_i}(f,\vect{x}_i,t_j)$ is the prediction made by a model trained on all data except those from location $\vect{x}_i$. Computing this error naively would require training $N_\text{s}$ separate models, leading to prohibitive computational costs. Instead, we develop an efficient approach that requires only a single matrix inversion. The prediction at location $\vect{x}_i$ based on data from all other locations can be expressed as:
\begin{equation}
  \begin{aligned}
  &\hat{\vect{q}}_{-\vect{x}_i}(\vect{x}_i, t_j)\\
  &= \big( \mathcal{K}_\text{f} 
  \otimes \mathcal{K}_\text{s}(\vect{x}_i, \mathcal{X}_{\text{tr}(-i)}) 
  \otimes \mathcal{K}_\text{t}(t_j, \mathcal{T}_\text{tr}) \big) \\
  &\quad \cdot \Big[ 
  \Sigma_{\text{tr,f}} \otimes \Sigma_{\text{tr}(-i),\text{s}} \otimes \Sigma_{\text{tr,t}} 
  + D \otimes I_{N_\text{s}(-i)} \otimes I_{N_\text{t}} 
  \Big]^{-1}\\
  &\quad \cdot\vect{y}_{\text{tr}(-i)}
  \label{eq:q-i}
\end{aligned}
\end{equation}
where $\mathcal{X}_{\text{tr}(-i)}=\{\vect{x}_1,\dots,\vect{x}_{i-1},\vect{x}_{i+1},\dots,\vect{x}_{N_\text{s}}\}$, $\Sigma_{\text{tr}(-i),\text{s}}=\mathcal{K}_\text{s}(\mathcal{X}_{\text{tr}(-i)},\mathcal{X}_{\text{tr}(-i)})$, and $\vect{y}_{\text{tr}(-i)}=[\vect{y}(\vect{x}_k,t_j)]_{\vect{x}_k\in \mathcal{X}_{\text{tr}(-i)},t_j\in\mathcal{T}_\text{tr}}$.

By partitioning the full covariance matrix and leveraging block matrix inversion formulas, we can express the leave-one-out prediction residuals in terms of the full model's inverse covariance matrix (see the Appendix for more details):
\begin{equation}
  \begin{aligned}
  &\left[ y(f, \vect{x}_i, t_j) 
  - \hat{q}_{-\vect{x}_i}(f, \vect{x}_i, t_j) \right]_{f \in F,\; t_j \in \mathcal{T}_\text{tr}}   \\
  &= \left( \hat{\vect{\delta}}^{*} \right)^{2}_{N_\text{f},\; 1_i,\; N_\text{t}} 
  \cdot \left( \Sigma_{\text{tr}}^{-1} \vect{y}_{\text{tr}} \right)_{N_\text{f},\; (1_i,\, \cdot),\; N_\text{t}}
  \label{eq:cross-val}
      \end{aligned}
\end{equation}
Hence, we only need to compute $\Sigma_{\text{tr}}^{-1}$ once. The calculation process is similar to Eq.~\eqref{eq:posterior-distribution}:
\begin{equation}
  \begin{aligned}
  \Sigma_{\text{tr}}^{-1} \vect{y}_{\text{tr}} 
  &= U \Lambda^{-1} U^\top \vect{y}_{\text{tr}}   \\[0.3em]
  &= 
  \begin{bmatrix}
  U_{\text{s}} \otimes U_{\text{t}} & 0 \\
  0 & U_{\text{s}} \otimes U_{\text{t}}
  \end{bmatrix}
  \begin{bmatrix}
  \tilde{\Lambda}_{11} & \tilde{\Lambda}_{12} \\
  \tilde{\Lambda}_{21} & \tilde{\Lambda}_{22}
  \end{bmatrix}
  \begin{bmatrix}
  \vect{v}_u \\
  \vect{v}_v
  \end{bmatrix}
    \\[0.3em]
  &= 
  \begin{bmatrix}
  \operatorname{vec} \left( U_{\text{t}}\, 
  \operatorname{vec}^{-1}(\tilde{\Lambda}_{11} \vect{v}_u + \tilde{\Lambda}_{12} \vect{v}_v)\, 
  U_{\text{s}}^\top \right) \\
  \operatorname{vec} \left( U_{\text{t}}\, 
  \operatorname{vec}^{-1}(\tilde{\Lambda}_{21} \vect{v}_u + \tilde{\Lambda}_{22} \vect{v}_v)\, 
  U_{\text{s}}^\top \right)
  \end{bmatrix}
  \label{eq:cro-val-err-f2-whole}
      \end{aligned}
\end{equation}
  
The element corresponding to the $i$-th spatial location is then extracted from Eq. (\ref{eq:cro-val-err-f2-whole}) as:
\begin{equation}
  \begin{aligned}
  &\left( \Sigma_{\text{tr}}^{-1} \mathbf{y}_{\text{tr}} \right)_{N_\text{f},\,(1_i,\, \cdot),\, N_\text{t}}   \\
  &= 
  \begin{bmatrix}
  \operatorname{vec} \left( 
  U_{\text{t}}\, 
  \operatorname{vec}^{-1} \left( 
  \tilde{\Lambda}_{11} \vect{v}_u + \tilde{\Lambda}_{12} \vect{v}_v 
  \right)\, 
  U_{\text{s}}^\top 
  \right)_{1_i,\, N_\text{t}} 
  \\[0.5em]
  \operatorname{vec} \left( 
  U_{\text{t}}\, 
  \operatorname{vec}^{-1} \left( 
  \tilde{\Lambda}_{21} \vect{v}_u + \tilde{\Lambda}_{22} \vect{v}_v 
  \right)\, 
  U_{\text{s}}^\top 
  \right)_{1_i,\, N_\text{t}} 
  \end{bmatrix}
  \label{eq:cro-val-err-f2-i}
      \end{aligned}
\end{equation}
where the first component corresponds to task-1 ($u$) and the second component to task-2 ($v$) in our two-task formulation. Eq. (\ref{eq:cro-val-err-f2-i}) is inserted back into Eq. (\ref{eq:cross-val}) to compute the corresponding leave-one-out residual, allowing us to efficiently compute cross-validation errors.

\section{Numerical Experiments} \label{s:results}
     
We assess the performance of our P-M-GP framework in predictive modeling of cardiac electrodynamics within a 3D ventricular geometry. The geometry is discretized into 1,094 nodes and 2,184 mesh elements, constituting a refined mesh derived from the geometry data in the 2007 PhysioNet Computing in Cardiology Challenge\cite{goldberger2000physiobank}. The simulation data is generated by numerically solving the FitzHugh-Nagumo (FHN) model using finite element methods. The reaction kinetics in the FHN model are specified as follows:
   \begin{equation}
      \begin{aligned}
      g_1(u,v) &= C_1 u (u - \alpha)(1 - u) + {\color{red}-} C_2 u v  \\
      g_2(u,v) &= b (u - d v)
  \end{aligned}
\end{equation}
 with FHN model parameters set to: $C_1 = 0.26$, $C_2 = 0.1$, $\alpha = 0.13$, $b = 0.013$, and $d = 1.0$; and diffusion parameters set to: $e_1=10$ and $e_2=0$. We applied two protocols to generate the simulation data: (1) Protocol I -- A regular-pacing activation source is placed at the apex of the ventricular geometry to stimulate the reaction-diffusion electrodynamics; (2) Protocol II -- An additional activation source is introduced to imitate self-sustained, disorganized dynamics under fibrillation. 

We denote the resulting simulation data as $\vect{q}(\vect{x}, t)=[\vect{u}(\vect{x}, t),\vect{v}(\vect{x}, t)]=[u(\vect{x}_i, t_j),v(\vect{x}_i, t_j)]_{i\in \mathcal{X} ,j\in \mathcal{T}}$ with $|\mathcal{X}|=1094$. Note that the time series signals collected at each spatial location $\vect{x}_i$ consist of 1570 data points for Protocol I (i.e., $|\mathcal{T}|=1570$), and 2920 points for Protocol II (i.e., $|\mathcal{T}|=2920$). Because measurement noise is inevitable in real-world data collection, we add different levels of noise to the simulation data to investigate the prediction performance. Specifically, the physical measurements are generated as
    $
    \vect{y}(\vect{x}, t) = \vect{q}(\vect{x}, t) + \vect{\xi}(\vect{x}, t)
    $,
    where $\vect{\xi}(\vect{x}, t)$ is the noise that follows a Gaussian distribution, $\vect{\xi}(\vect{x}, t) \sim \sigma_\xi \cdot \mathcal{N}(\vect{0}, 1)$, where $\sigma_\xi$ is the noise level coefficient. Our P-M-GP is compared with the traditional M-GP based on the relative error ($RE$):
    \begin{equation}
      \begin{aligned}
        RE &= \frac{\| \vect{\hat{q}}(\vect{x}, t) - \vect{q}(\vect{x}, t) \|}{ \| \vect{q}(\vect{x}, t) \|}
      \end{aligned}
    \end{equation}
    where $\vect{q}(\vect{x}, t)$ and $\vect{\hat{q}}(\vect{x}, t)$ denote the reference and predicted cardiac dynamics, respectively.

\subsection{Prediction under Simulation Protocol I }


\begin{figure*}
	\centering
	\includegraphics[width=1.0\linewidth]{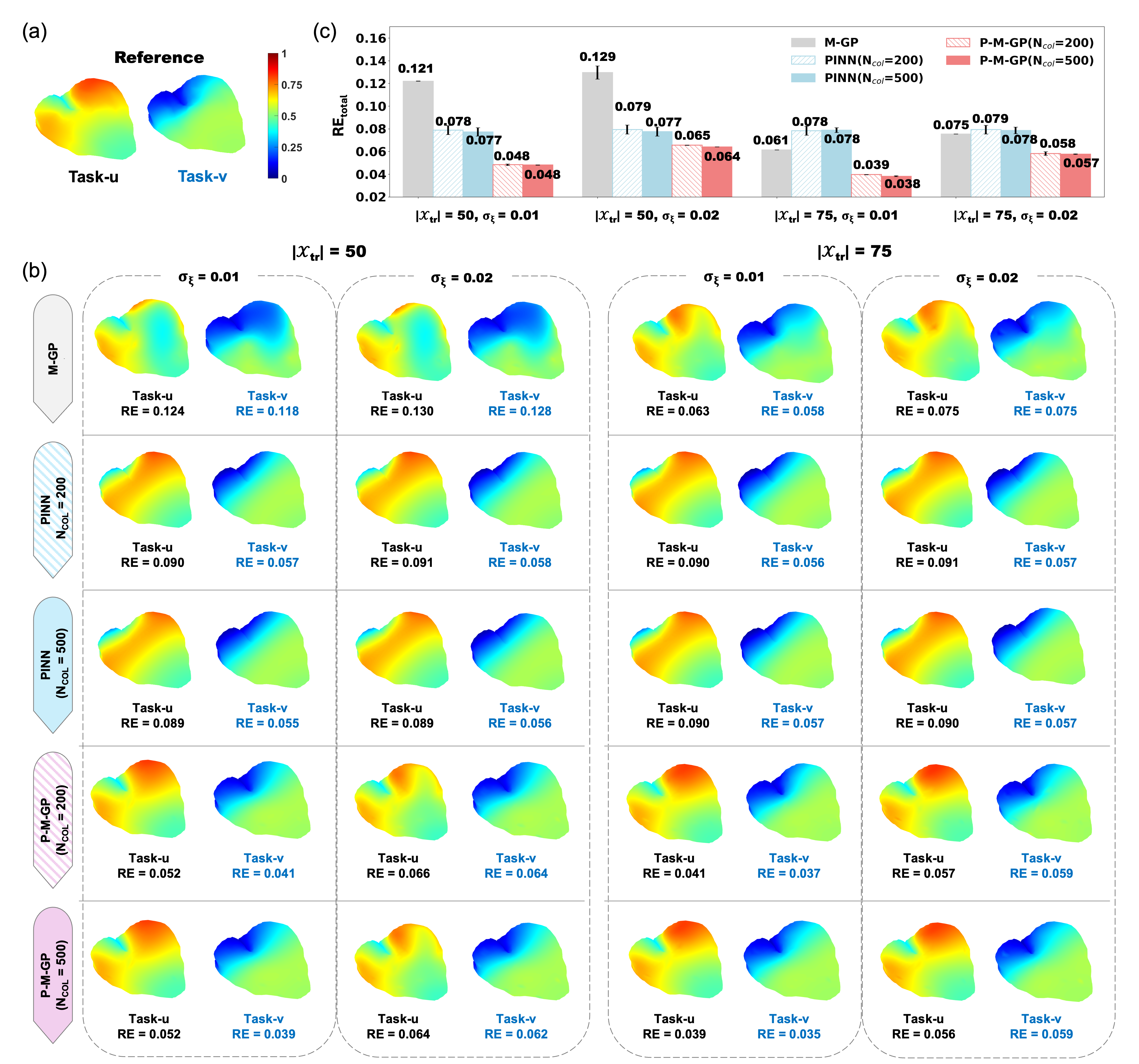}
    \caption{Prediction results under Simulation Protocol I: (a) Reference mapping of task-u and task-v under Protocol I at time point $t=800$. (b) Estimated mappings by M-GP, PINN ($N_\text{col}=200$ or $500$) and our P-M-GP model ($N_\text{col}=200$ or $500$) under different training dataset sizes ($|\mathcal{X}_\text{tr}| = 50, 75$) and noise levels ($\sigma_\xi = 0.01, 0.02$) at time $t=800$. (c) Bar chart comparing the aggregated $RE$ from 3 replications.}
	\label{Fig:Pred-vis_Hea}
\end{figure*}

Fig.~\ref{Fig:Pred-vis_Hea} compares the spatiotemporal prediction accuracy of our P-M-GP model with that of M-GP and PINN, across two training set sizes (\(|\mathcal{X}_\text{tr}| = 50, 75\)) and two noise levels (\(\sigma_\xi = 0.01, 0.02\)). Panel (a) shows the ground-truth spatial distribution of cardiac electrodynamics at time step \(t = 800\). Panel (b) displays the reconstructed mappings for both variables (task-u: normalized electric potential; task-v: recovery electric current) produced by each method. Across all experimental settings, P-M-GP consistently achieves higher fidelity to the reference spatial patterns shown in Fig.~\ref{Fig:Pred-vis_Hea}(a), whereas the predictions from M-GP (which lacks physics-based constraints) and standard PINN (which does not leverage geometric manifold features) show more pronounced deviations from the ground truth.


Fig.~\ref{Fig:Pred-vis_Hea}(c) presents the quantitative performance comparison based on the aggregated $RE$ metric ($RE_{\text{total}}=\frac{1}{2}(RE_u+RE_v)$), calculated from triplicate experiments with randomized seeds. The performance is reported as the mean $\pm$ standard deviation, highlighting the superior prediction accuracy achieved by the P-M-GP model. For \(N_\text{col}=200\) and \(|\mathcal{X}_\text{tr}|=50\), our P-M-GP method yields $RE$ values of $0.048 \pm (1.58\times10^{-3})$ when  \(\sigma_\xi=0.01\) and $0.065 \pm (1.66\times10^{-3})$ when \(\sigma_\xi=0.02\), achieving the $RE$ reduction of \(60.33\%\) and \(49.61\%\) compared to the baseline M-GP, and improvements of $38.46\%$ and $17.72\%$ over the standard PINN, respectively. 

When the training size increases to \(|\mathcal{X}_\text{tr}|=75\), the performance gains achieved by P-M-GP is also prominent: $RE$ is further reduced to $0.039 \pm (9.38\times10^{-4})$ for  \(\sigma_\xi=0.01\) and $0.058 \pm (8.08\times10^{-4})$ for \(\sigma_\xi=0.02\), corresponding to reductions of \(36.06\%\)  and \(22.66\%\) compared to M-GP, and $50.00\%$ and $26.58\%$ over PINN, respectively. Notably, as the dataset size increases, M-GP demonstrates a marked performance advantage over PINN. This performance gap is particularly significant given the substantial difference in model complexity: the PINN framework adopted in our comparison is a feedforward neural network with six hidden dense layers and 515 trainable parameters, while the M-GP model is highly compact with only 9 trainable parameters. Despite PINN's greater parameterization and capacity, its predictive accuracy remains inferior to both M-GP and P-M-GP, further highlighting the efficiency and effectiveness of geometry-aware GP framework.

A similar pattern is observed when the number of collocation points increases to \(N_\text{col}=500\): $RE$ reductions of \(60.33\%\) (\(\sigma_\xi=0.01\)) and \(50.38\%\) (\(\sigma_\xi=0.02\)) at \(|\mathcal{X}_\text{tr}|=50\), and \(37.70\%\) and \(24.00\%\) at \(|\mathcal{X}_\text{tr}|=75\) compared to M-GP. Compared to PINN, the $RE$ reductions are \(37.66\%\) (\(\sigma_\xi=0.01,\, |\mathcal{X}_\text{tr}|=50\)), \(16.88\%\) (\(\sigma_\xi=0.02,\, |\mathcal{X}_\text{tr}|=50\)), \(51.28\%\) (\(\sigma_\xi=0.01,\, |\mathcal{X}_\text{tr}|=75\)), and \(26.92\%\) (\(\sigma_\xi=0.02,\, |\mathcal{X}_\text{tr}|=75\)), respectively.
It is worth noting that increasing the number of collocation points \(N_\text{col}\) 
from 200 to 500 yields marginal improvements in $RE$ for both PINN and P-M-GP. This indicates a possible saturation effect, where the additional collocation points provide diminishing returns in encoding the underlying physical constraints. The limited impact is likely due to the already sufficient physical information captured by \(N_\text{col}=200\), suggesting that beyond a certain threshold, further increasing the density of physics-based supervision does not significantly enhance the model performance.

\begin{figure*}
	\centering
	\includegraphics[width=1\linewidth]{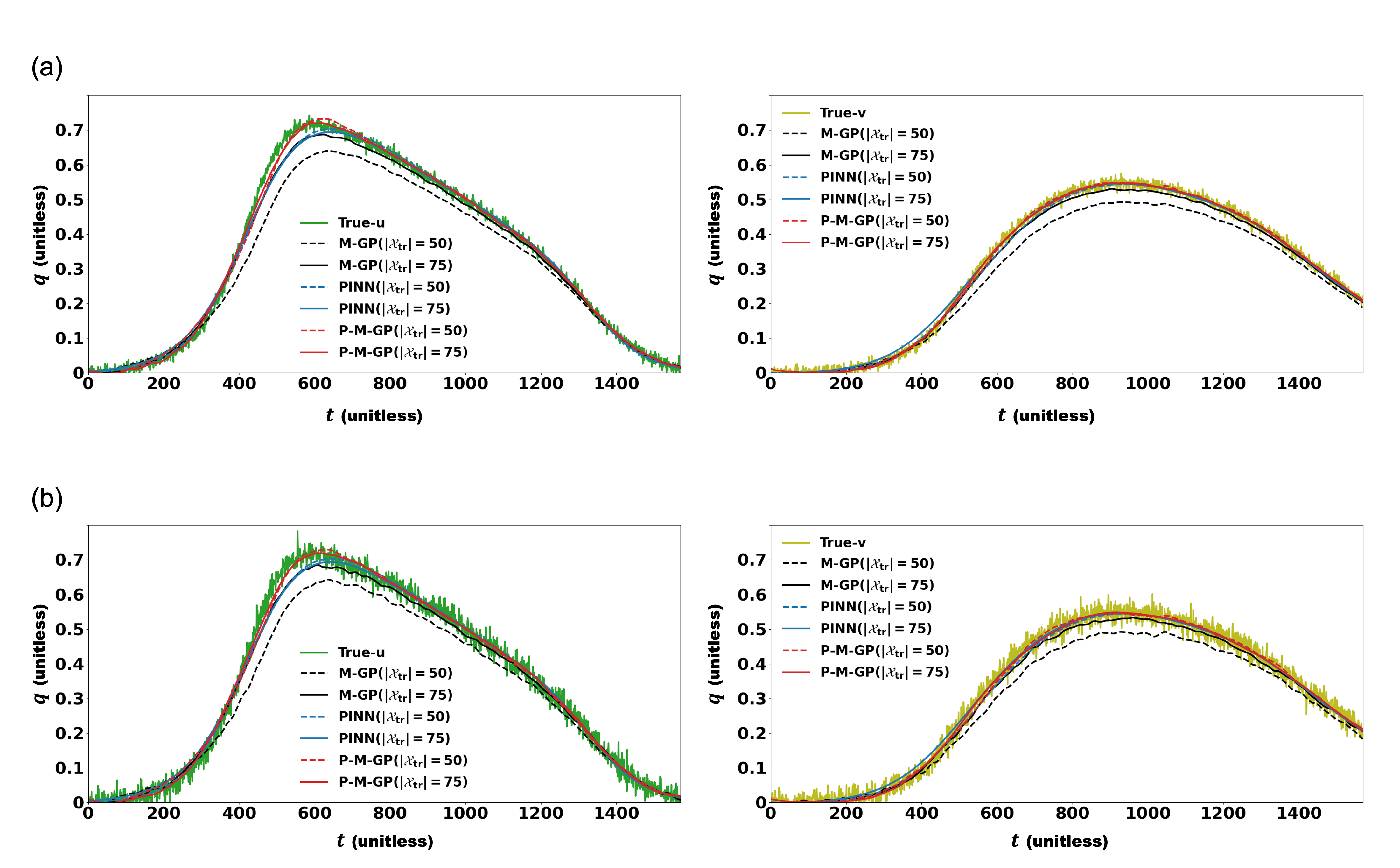}
    \caption{Temporal evolution of cardiac electrodynamics at spatial location \(\vect{x}_{599}\) under Protocol I. The predictions from M-GP, PINN ($N_\text{col}=500$) and P-M-GP ($N_\text{col}=500$) are compared under varying training data sizes ($|\mathcal{X}_\text{tr}| = 50, 75$) and noise levels: (a) $\sigma_\xi = 0.01$, (b) $\sigma_\xi = 0.02$.}
	\label{Fig:Pred-evol_Hea}
\end{figure*} 

Fig.~\ref{Fig:Pred-evol_Hea} illustrates the temporal evolution of the normalized electric potential 
(task-u) and the recovery electric current (task-v) at spatial location \(\vect{x}_{599}\), comparing the 
predictive performance of the proposed P-M-GP model against  M-GP and PINN under two training set sizes: \(|\mathcal{X}_\text{tr}| = 50\) and \(|\mathcal{X}_\text{tr}| = 75\). When 
trained with \(|\mathcal{X}_\text{tr}| = 50\), P-M-GP produces evolution curves overlapping with the 
ground-truth physical observations for both noise levels, demonstrating high predictive accuracy across 
both tasks. In contrast, M-GP displays noticeable deviations from the reference, while PINN achieves improved accuracy over M-GP but still exhibits clear discrepancies. As the training set increases to \(|\mathcal{X}_\text{tr}| = 75\), P-M-GP maintains its superior estimation fidelity. This improvement is primarily due to the incorporation of physics-augmented constraints, which enable the model to better capture and preserve the underlying physical dynamics. In comparison, the traditional M-GP, being purely data-driven, lacks the ability to encode governing physical laws, and PINN is less effective at exploiting non-Euclidean spatial structures, leading to less accurate and less stable predictions, particularly when training data is limited.

\subsection{Prediction under Simulation Protocol II}



\begin{figure*}
	\centering
	\includegraphics[width=1.0\linewidth]{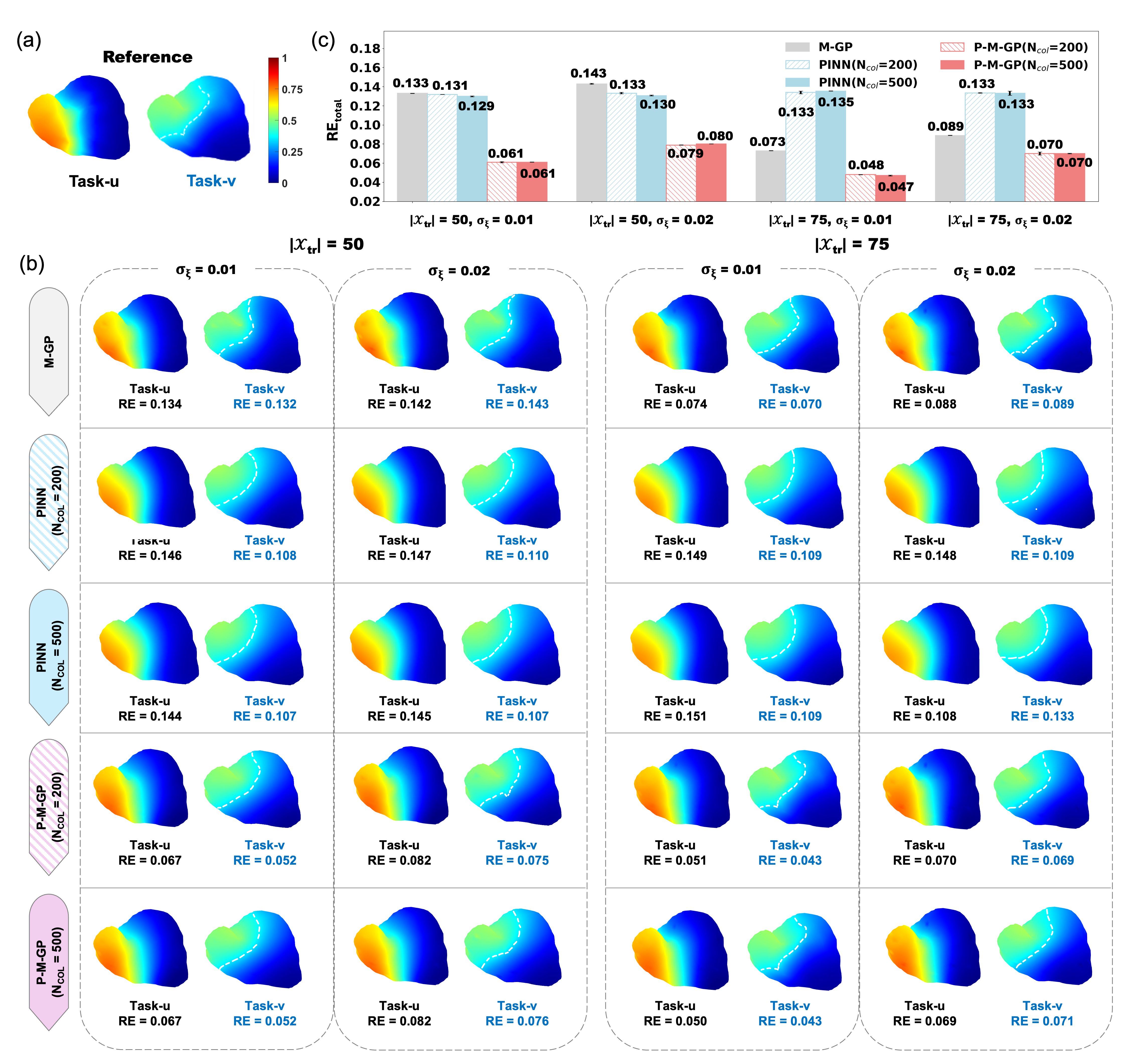}
    \caption{Prediction results under Protocol II: (a) Ground-truth visualization at $t=800$ under Simulation Protocol II. (b) Predicted distributions from M-GP, PINN ($N_\text{col}=200$ or $500$) and P-M-GP ($N_\text{col}=200$ or $500$) using varying training sizes ($|\mathcal{X}_\text{tr}| = 50, 75$) and noise conditions ($\sigma_\xi = 0.01, 0.02$) at time $t=800$. White dotted lines are marked for better visualization. (c) Comparison of the aggregated $RE$ between P-M-GP and M-GP averaged over 3 trials.}
	\label{Fig:Pred-vis_Dis}
\end{figure*} 

Fig.~\ref{Fig:Pred-vis_Dis} shows the comparison  across M-GP, PINN and our P-M-GP with training set sizes \(|\mathcal{X}_\text{tr}| = 50, 75\) and noise levels \(\sigma_\xi = 0.01, 0.02\) under Simulation Protocol II. Panel (a) shows the ground-truth electrodynamics at time \(t = 800\). Reconstructed mappings in Panel (b) reveal that P-M-GP continues to significantly outperform M-GP and PINN across both tasks. In particular, at \(|\mathcal{X}_\text{tr}|=50\) and \(N_\text{col}=500\), the $RE$ reductions achieved by P-M-GP for task-u are \(49.25\%\) (\(\sigma_\xi=0.01\)) and \(42.25\%\) (\(\sigma_\xi=0.02\)), with corresponding improvements of \(60.60\%\) and \(46.85\%\) for task-v respectively, compared to the baseline M-GP. Compared to PINN, the improvements are 53.47\% (\(\sigma_\xi=0.01\)) and 43.44\% (\(\sigma_\xi=0.02\)) for task-u, while 51.40\% (\(\sigma_\xi=0.01\)) and 28.97\% (\(\sigma_\xi=0.02\)) for task-v.


  Fig.~\ref{Fig:Pred-vis_Dis}(c) presents the aggregated $RE$ ($RE_{\text{total}}$) averaged over three independent trials with randomized seeds to generate data noise. For \(N_\text{col} = 200\) and \(|\mathcal{X}_\text{tr}| = 50\), the P-M-GP model attains RE values of \(0.061 \pm (7.39 \times 10^{-4})\) for \(\sigma_\xi = 0.01\) and \(0.079 \pm (1.51 \times 10^{-4})\) for \(\sigma_\xi = 0.02\), corresponding to substantial reductions of \(54.13\%\) and \(44.75\%\) compared to the M-GP baseline. Compared to PINN (\(N_\text{col} = 200\)), the improvements are 53.43\% and 40.60\%, respectively. When the training set increases to \(|\mathcal{X}_\text{tr}| = 75\), the RE values further decrease to \(0.048 \pm (1.05 \times 10^{-4})\) for \(\sigma_\xi = 0.01\) and \(0.070 \pm (1.64 \times 10^{-3})\) for \(\sigma_\xi = 0.02\), yielding relative improvements of \(34.24\%\) and \(21.34\%\) over M-GP. Compared to PINN, $RE$ is reduced by 63.90\% and 47.36\%, with $RE$ values of \(0.133 \pm (1.43 \times 10^{-3})\) and \(0.133 \pm (4.16 \times 10^{-4})\) for PINN, respectively. Consistent with Simulation Protocol I, increasing the number of collocation points from \(N_\text{col} = 200\) to 500 results in only marginal additional reductions in $RE$, indicating a possible saturation effect where further increases in collocation points yield diminishing returns in capturing the underlying physics.

\begin{figure*}
	\centering
	\includegraphics[width=1.0\linewidth]{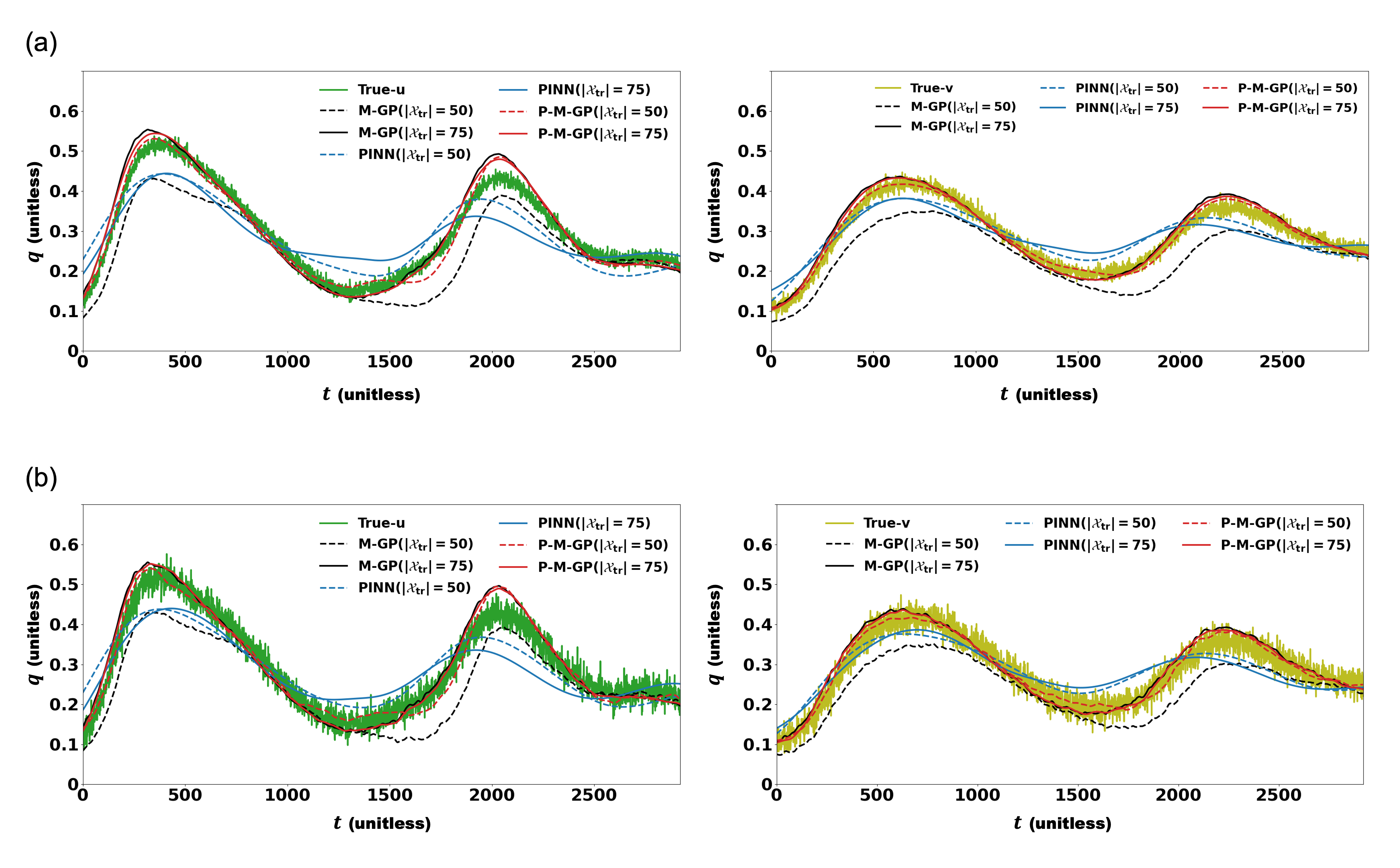}
    \caption{Comparison of temporal dynamics under Protocol II at spatial location \(\vect{x}_{636}\) between M-GP, PINN ($N_\text{col}=500$) and P-M-GP ($N_\text{col}=500$) predictions across different training sample sizes ($|\mathcal{X}_\text{tr}|=50, 75$) and noise conditions: (a) $\sigma_\xi = 0.01$ and (b) $\sigma_\xi = 0.02$.}
	\label{Fig:Pred-evol_Dis}
\end{figure*}

  Temporal dynamics at spatial node \(\vect{x}_{636}\) under Simulation Protocol II are illustrated in Fig.~\ref{Fig:Pred-evol_Dis}. Compared to the waveforms observed in Simulation Protocol I, the dynamics here exhibit greater irregularity, reflecting the more complex nature of pathological cardiac conditions. Despite these challenges, the proposed P-M-GP model demonstrates superior predictive performance. It accurately captures the temporal evolution of both task-\(u\) (green) and task-\(v\) (olive) variables, maintaining close alignment with the ground truth. 
  In contrast, the M-GP baseline displays clear amplitude distortions and temporal lag, particularly when the training set is limited (\(|\mathcal{X}_\text{tr}| = 50\)). The PINN model exhibits even greater discrepancies from the ground truth. These results collectively highlight the robustness of the proposed physics-augmented approach in predictive modeling of complex and chaotic spatiotemporal dynamics.


\section{Conclusions} \label{s:conclusions}

This paper presents a physics-augmented multi-task Gaussian Process (P-M-GP) framework for predictive modeling of spatiotemporal dynamic systems. The proposed methodology integrates geometric awareness, task-wise dependency modeling, and physics-informed regularization within a unified GP framework, addressing key challenges associated with irregular spatial geometries, complex temporal dynamics, and the need for multi-output prediction. Specifically, we extend the geometry-aware spatiotemporal GP (G-ST-GP) framework with an M-GP model to jointly capture dependencies between interconnected tasks in spatiotemporal systems. Additionally, by exploiting the Kronecker product structure of the kernel, our framework achieves computational efficiency in posterior predictions, enabling scalable modeling of high-dimensional systems.  Furthermore, we develop a physics-augmented learning strategy that incorporates domain knowledge through physics-based regularization within the M-GP. This approach ensures that predictions adhere to underlying physical principles while maintaining data-driven flexibility, addressing the limitations of purely data-driven methods in sparse data scenarios. We validate our P-M-GP framework through numerical experiments in 3D cardiac electrophysiological modeling. Results demonstrate that our method significantly outperforms data-driven-only and geometry-prior-agnostic approaches. By bridging statistical learning with domain-specific physics and geometric priors, our framework offers an effective solution for complex spatiotemporal modeling tasks, with broad applicability to real-world dynamic systems.


\section*{Appendix}
\subsection{Details of Eq.~\eqref{eq:dK_dt}}
 Under the Matérn temporal kernel with smoothness parameter $\nu=3/2$, we have:
 \begin{equation}
  \begin{aligned}
  &\left[ \frac{\partial \mathcal{K}_\text{t}^*}{\partial t} \right]_{i,j}\\
  &= \frac{\partial}{\partial t_i} 
  \left\{ \sigma_a \left( 1 + \frac{\sqrt{3} |t_i - t_j|}{l_t} \right) 
  \exp \left( -\frac{\sqrt{3} |t_i - t_j|}{l_t} \right) \right\}\\
  &= - \sigma_a \cdot \frac{3}{l_t^2} |t_i - t_j| 
  \exp \left( -\frac{\sqrt{3} |t_i - t_j|}{l_t} \right) 
  \cdot \operatorname{sign}(t_i - t_j)
  \label{eq:dKdt-matern32}
      \end{aligned}
\end{equation}
with definitions of $m = \|t_i - t_j\|$ and $a = \frac{\sqrt{3}}{l_t}$, it becomes:
\begin{equation}
  \begin{aligned}
  &\left[ \frac{\partial \mathcal{K}_t^*}{\partial t} \right]_{i,j}
  = \frac{\partial}{\partial t_i} 
  \left[ \sigma_a (1 + a m) \exp(-a m) \right] \\
  &= \sigma_a [ a \frac{\partial m}{\partial t_i} \exp(-a m) 
  - a (1 + a m) \\[0.3em]
  &\exp(-a m) \frac{\partial m}{\partial t_i} ]  \\
  &= -\sigma_a a^2 m \exp(-a m) \cdot \frac{\partial m}{\partial t_i} \\
  &= -\sigma_a a^2 m \exp(-a m) \cdot \operatorname{sign}(t_i - t_j)
  \label{eq:dt_matern32_stepwise}
      \end{aligned}
\end{equation} 
where
\begin{equation}
  \frac{\partial m}{\partial t_i} = \frac{\partial \| t_i - t_j \|}{\partial t_i} = \operatorname{sign} (t_i - t_j) = 
  \begin{cases}
      1, & t_i > t_j \\
      0, & t_i = t_j \\
      -1, & t_i < t_j
  \end{cases}
\end{equation}

\subsection{Details of Eq.~\eqref{eq:cross-val}}

If we define $\Sigma_{\text{tr}(-i)}=\Sigma_{\text{tr,f}} \otimes \Sigma_{\text{tr(-i),s}}\otimes\Sigma_{\text{tr,t}} + D \otimes I_{N_s(-i)} \otimes I_{N_\text{t}}$, to isolate the dynamic signals collected from the $i^\text{th}$ spatial location, we partition $\Sigma_{\text{tr}}$ into block matrices:
    \begin{equation}
        \Sigma_{\text{tr}} = \begin{pmatrix}
                \Sigma_{\text{tr}(-i)} & \mathcal{K}_{(1_i, N_s - 1_i)}^\top \\
                \mathcal{K}_{(1_i, N_s - 1_i)} & \mathcal{K}_{(1_i, 1_i)}
            \end{pmatrix}
        \label{eq:sigma_block}
    \end{equation}
    where $\mathcal{K}_{(1_i, N_s - 1_i)} = \mathcal{K}_{\text{f}}(F, F) \otimes \mathcal{K}_\text{s}(\vect{x}_i,\mathcal{X}_{\text{tr}(-i)}) \otimes \mathcal{K}_\text{t}(\mathcal{T}_\text{tr},\mathcal{T}_\text{tr}) + D \otimes I_{(1_i, N_s - 1_i)} \otimes I_{N_\text{t}}$ is the covariance between the dynamics at the $i^\text{th}$ location and the rest of the training data, and $\mathcal{K}_{(1_i, 1_i)} = \mathcal{K}_{\text{f}}(F, F) \otimes \mathcal{K}_\text{s}(\vect{x}_i,\vect{x}_i) \otimes \mathcal{K}_\text{t}(\mathcal{T}_\text{tr},\mathcal{T}_\text{tr}) + D \otimes I_{\text{s},(1_i, 1_i)} \otimes I_{N_\text{t}}$ is the covariance matrix for the dynamics at the $i^\text{th}$ location. Using the block matrix inversion formula, we have:
    \begin{equation}
        \begin{aligned}
            \Sigma_{\text{tr}} ^{-1} &:= \begin{pmatrix}
                A_{11} & A_{21}^\top \\
                A_{21} & A_{22}
            \end{pmatrix}        
        \end{aligned}
        \label{eq:inv-block}
    \end{equation}
    where $A_{21} = -A_{22} \mathcal{K}_{(1_i, N_s - 1_i)} \Sigma_{\text{tr}(-i)}^{-1}$ and $A_{22}^{-1} = \left[\left( \Sigma_{\text{tr}}^{-1} \right)_{(1_i, 1_i)}\right]^{-1}=\mathcal{K}_{(1_i, 1_i)}-\mathcal{K}_{(1_i, N_s - 1_i)} \Sigma_{\text{tr}(-i)}^{-1} \mathcal{K}_{(1_i, N_s - 1_i)}^\top$ is the estimated variance of location $i$, i.e. $(\hat{\vect{\delta}}^{*})^{2}_{N_\text{f}, 1_i, N_\text{t}}$.
Hence, we can simplify Eq.~\eqref{eq:q-i} and calculate the residual as:
    \begin{equation}
      \begin{aligned}
          &\left[y(f,\vect{x}_i,t_j) - \hat{q}_{-\vect{x}_i}(f,\vect{x}_i,t_j)\right]_{f\in F, t_j\in \mathcal{T}_\text{tr}} \\
          &= A_{22}^{-1} \left( A_{22} \vect{y}_{(i)} -A_{22} \mathcal{K}_{(1_i, N_s - 1_i)} \Sigma_{\text{tr}(-i)}^{-1}  \vect{y}_{\text{tr}(-i)} \right)\\
          &= A_{22}^{-1} \left( A_{22} \vect{y}_{(i)} + A_{21} \vect{y}_{\text{tr}(-i)} \right)\\
          &= (\hat{\vect{\delta}}^{*})^{2}_{N_\text{f}, 1_i, N_\text{t}} \left( \Sigma_{\text{tr}}^{-1} \vect{y}_{\text{tr}} \right)_{N_\text{f},(1_i, \cdot),N_\text{t}}
      \end{aligned}
      \label{eq:residual_implified}
  \end{equation}
where $\vect{y}_{(i)}=\left[y(f, x_i, t_j)\right]_{f\in F, t_j\in \mathcal{T}_\text{tr}} $, and $\left( \Sigma_{\text{tr}}^{-1} \mathbf{y}_{\text{tr}} \right)_{N_\text{f},(1_i, \cdot),N_\text{t}}$ denotes the sub-vector of $ \Sigma_{\text{tr}}^{-1} \mathbf{y}_{\text{tr}} $ corresponding to the $i$-th spatial location.

{
\bibliographystyle{IEEEtran}
\bibliography{Ref}

\begin{thebibliography}{10}
\providecommand{\url}[1]{#1}
\csname url@samestyle\endcsname
\providecommand{\newblock}{\relax}
\providecommand{\bibinfo}[2]{#2}
\providecommand{\BIBentrySTDinterwordspacing}{\spaceskip=0pt\relax}
\providecommand{\BIBentryALTinterwordstretchfactor}{4}
\providecommand{\BIBentryALTinterwordspacing}{\spaceskip=\fontdimen2\font plus
\BIBentryALTinterwordstretchfactor\fontdimen3\font minus \fontdimen4\font\relax}
\providecommand{\BIBforeignlanguage}[2]{{%
\expandafter\ifx\csname l@#1\endcsname\relax
\typeout{** WARNING: IEEEtran.bst: No hyphenation pattern has been}%
\typeout{** loaded for the language `#1'. Using the pattern for}%
\typeout{** the default language instead.}%
\else
\language=\csname l@#1\endcsname
\fi
#2}}
\providecommand{\BIBdecl}{\relax}
\BIBdecl

\bibitem{stroud2001dynamic}
J.~R. Stroud, P.~M{\"u}ller, and B.~Sans{\'o}, ``Dynamic models for spatiotemporal data,'' \emph{Journal of the Royal Statistical Society: Series B (Statistical Methodology)}, vol.~63, no.~4, pp. 673--689, 2001.

\bibitem{yang2023sensing}
H.~Yang and B.~Yao, \emph{Sensing, Modeling and Optimization of Cardiac Systems: A New Generation of Digital Twin for Heart Health Informatics}.\hskip 1em plus 0.5em minus 0.4em\relax Springer Nature, 2023.

\bibitem{trayanova2011whole}
N.~A. Trayanova, ``Whole-heart modeling: applications to cardiac electrophysiology and electromechanics,'' \emph{Circulation research}, vol. 108, no.~1, pp. 113--128, 2011.

\bibitem{sim2020epicardial}
K.~Sim, F.~Ershad, Y.~Zhang, P.~Yang, H.~Shim, Z.~Rao, Y.~Lu, A.~Thukral, A.~Elgalad, Y.~Xi \emph{et~al.}, ``An epicardial bioelectronic patch made from soft rubbery materials and capable of spatiotemporal mapping of electrophysiological activity,'' \emph{Nature Electronics}, vol.~3, no.~12, pp. 775--784, 2020.

\bibitem{yao2018multifractal}
B.~Yao, F.~Imani, A.~S. Sakpal, E.~Reutzel, and H.~Yang, ``Multifractal analysis of image profiles for the characterization and detection of defects in additive manufacturing,'' \emph{Journal of Manufacturing Science and Engineering}, vol. 140, no.~3, p. 031014, 2018.

\bibitem{wang2024sub}
R.~Wang, R.~Wang, C.~Dou, S.~Yang, R.~Gnanasambandam, A.~Wang, and Z.~Kong, ``Sub-surface thermal measurement in additive manufacturing via machine learning-enabled high-resolution fiber optic sensing,'' \emph{Nature Communications}, vol.~15, no.~1, p. 7568, 2024.

\bibitem{xie2024effect}
J.~Xie, B.~Yao, and Z.~Jiang, ``The effect of different optimization strategies to physics-constrained deep learning for soil moisture estimation,'' \emph{arXiv preprint arXiv:2403.08154}, 2024.

\bibitem{thompson2015overview}
S.~M. Thompson, L.~Bian, N.~Shamsaei, and A.~Yadollahi, ``An overview of direct laser deposition for additive manufacturing; part i: Transport phenomena, modeling and diagnostics,'' \emph{Additive Manufacturing}, vol.~8, pp. 36--62, 2015.

\bibitem{yao2021constrained}
B.~Yao, Y.~Chen, and H.~Yang, ``Constrained markov decision process modeling for optimal sensing of cardiac events in mobile health,'' \emph{IEEE Transactions on Automation Science and Engineering}, vol.~19, no.~2, pp. 1017--1029, 2021.

\bibitem{wang2021knowledge}
L.~Wang, A.~Hawkins-Daarud, K.~R. Swanson, L.~S. Hu, and J.~Li, ``Knowledge-infused global-local data fusion for spatial predictive modeling in precision medicine,'' \emph{IEEE Transactions on Automation Science and Engineering}, vol.~19, no.~3, pp. 2203--2215, 2021.

\bibitem{yao2024multi}
B.~Yao, ``Multi-source data and knowledge fusion via deep learning for dynamical systems: applications to spatiotemporal cardiac modeling,'' \emph{IISE Transactions on Healthcare Systems Engineering}, pp. 1--14, 2024.

\bibitem{yao2020spatiotemporal}
B.~Yao and H.~Yang, ``Spatiotemporal regularization for inverse ecg modeling,'' \emph{IISE Transactions on Healthcare Systems Engineering}, vol.~11, no.~1, pp. 11--23, 2020.

\bibitem{yang2019internet}
H.~Yang, S.~Kumara, S.~T. Bukkapatnam, and F.~Tsung, ``The internet of things for smart manufacturing: A review,'' \emph{IISE transactions}, vol.~51, no.~11, pp. 1190--1216, 2019.

\bibitem{karniadakis2021physics}
G.~E. Karniadakis, I.~G. Kevrekidis, L.~Lu, P.~Perdikaris, S.~Wang, and L.~Yang, ``Physics-informed machine learning,'' \emph{Nature Reviews Physics}, vol.~3, no.~6, pp. 422--440, 2021.

\bibitem{guo2022deep}
S.~Guo, W.~Guo, L.~Bian, and Y.~Guo, ``A deep-learning-based surrogate model for thermal signature prediction in laser metal deposition,'' \emph{IEEE Transactions on Automation Science and Engineering}, vol.~20, no.~1, pp. 482--494, 2022.

\bibitem{ko2023framework}
H.~Ko, Y.~Lu, Z.~Yang, N.~Y. Ndiaye, and P.~Witherell, ``A framework driven by physics-guided machine learning for process-structure-property causal analytics in additive manufacturing,'' \emph{Journal of Manufacturing Systems}, vol.~67, pp. 213--228, 2023.

\bibitem{chen2016numerical}
Y.~Chen and H.~Yang, ``Numerical simulation and pattern characterization of nonlinear spatiotemporal dynamics on fractal surfaces for the whole-heart modeling applications,'' \emph{The European Physical Journal B}, vol.~89, pp. 1--16, 2016.

\bibitem{yao2016mesh}
B.~Yao, S.~Pei, and H.~Yang, ``Mesh resolution impacts the accuracy of inverse and forward ecg problems,'' in \emph{2016 38th Annual International Conference of the IEEE Engineering in Medicine and Biology Society (EMBC)}.\hskip 1em plus 0.5em minus 0.4em\relax IEEE, 2016, pp. 4047--4050.

\bibitem{calandra2016manifold}
R.~Calandra, J.~Peters, C.~E. Rasmussen, and M.~P. Deisenroth, ``Manifold gaussian processes for regression,'' in \emph{2016 International joint conference on neural networks (IJCNN)}.\hskip 1em plus 0.5em minus 0.4em\relax IEEE, 2016, pp. 3338--3345.

\bibitem{xie2024kronecker}
J.~Xie and B.~Yao, ``Hierarchical active learning for defect localization in 3d systems,'' \emph{IISE Transactions on Healthcare Systems Engineering}, vol.~14, no.~2, pp. 115--129, 2024.

\bibitem{clayton2011models}
R.~Clayton, O.~Bernus, E.~Cherry, H.~Dierckx, F.~H. Fenton, L.~Mirabella, A.~V. Panfilov, F.~B. Sachse, G.~Seemann, and H.~Zhang, ``Models of cardiac tissue electrophysiology: progress, challenges and open questions,'' \emph{Progress in biophysics and molecular biology}, vol. 104, no. 1-3, pp. 22--48, 2011.

\bibitem{10824917}
B.~Yao, F.~Leonelli, and H.~Yang, ``Simulation optimization of spatiotemporal dynamics in 3d geometries,'' \emph{IEEE Transactions on Automation Science and Engineering}, vol.~22, pp. 10\,442--10\,456, 2025.

\bibitem{caruana1997multitask}
R.~Caruana, ``Multitask learning,'' \emph{Machine Learning}, vol.~28, no.~1, pp. 41--75, 1997.

\bibitem{yao2025geometry}
X.~Zhang and B.~Yao, ``Geometry-aware active learning of spatiotemporal dynamic systems,'' \emph{arXiv preprint arXiv:2504.19012}, 2025.

\bibitem{karpatne2017theory}
A.~Karpatne, G.~Atluri, J.~H. Faghmous, M.~Steinbach, A.~Banerjee, A.~Ganguly, S.~Shekhar, N.~Samatova, and V.~Kumar, ``Theory-guided data science: A new paradigm for scientific discovery from data,'' \emph{IEEE Transactions on knowledge and data engineering}, vol.~29, no.~10, pp. 2318--2331, 2017.

\bibitem{yan2018real}
H.~Yan, K.~Paynabar, and J.~Shi, ``Real-time monitoring of high-dimensional functional data streams via spatio-temporal smooth sparse decomposition,'' \emph{Technometrics}, vol.~60, no.~2, pp. 181--197, 2018.

\bibitem{guo2020hierarchical}
S.~Guo, W.~Guo, and L.~Bain, ``Hierarchical spatial-temporal modeling and monitoring of melt pool evolution in laser-based additive manufacturing,'' \emph{IISE Transactions}, vol.~52, no.~9, pp. 977--997, 2020.

\bibitem{liu2022statistical}
X.~Liu, K.~Yeo, and S.~Lu, ``Statistical modeling for spatio-temporal data from stochastic convection-diffusion processes,'' \emph{Journal of the American Statistical Association}, vol. 117, no. 539, pp. 1482--1499, 2022.

\bibitem{xie2022physics}
J.~Xie and B.~Yao, ``Physics-constrained deep learning for robust inverse ecg modeling,'' \emph{IEEE Transactions on Automation Science and Engineering}, 2022.

\bibitem{yao2016physics}
B.~Yao and H.~Yang, ``Physics-driven spatiotemporal regularization for high-dimensional predictive modeling: A novel approach to solve the inverse ecg problem,'' \emph{Scientific reports}, vol.~6, no.~1, p. 39012, 2016.

\bibitem{williams2006gaussian}
C.~K. Williams and C.~E. Rasmussen, \emph{Gaussian processes for machine learning}.\hskip 1em plus 0.5em minus 0.4em\relax MIT press Cambridge, MA, 2006, vol.~2, no.~3.

\bibitem{liu2020gaussian}
H.~Liu, Y.-S. Ong, X.~Shen, and J.~Cai, ``When gaussian process meets big data: A review of scalable gps,'' \emph{IEEE transactions on neural networks and learning systems}, vol.~31, no.~11, pp. 4405--4423, 2020.

\bibitem{hu2020gaussian}
Z.~Hu, D.~Du, and Y.~Du, ``Gaussian process-based spatiotemporal modeling of electrical wave propagation in human atrium,'' in \emph{2020 42nd Annual International Conference of the IEEE Engineering in Medicine \& Biology Society (EMBC)}.\hskip 1em plus 0.5em minus 0.4em\relax IEEE, 2020, pp. 2602--2605.

\bibitem{wang2018spatial}
R.~Wang, L.~Zhang, and N.~Chen, ``Spatial correlated data monitoring in semiconductor manufacturing using gaussian process model,'' \emph{IEEE Transactions on Semiconductor Manufacturing}, vol.~32, no.~1, pp. 104--111, 2018.

\bibitem{soh2014spatio}
H.~Soh and Y.~Demiris, ``Spatio-temporal learning with the online finite and infinite echo-state gaussian processes,'' \emph{IEEE transactions on neural networks and learning systems}, vol.~26, no.~3, pp. 522--536, 2014.

\bibitem{Senanayake_Ramos_2016}
\BIBentryALTinterwordspacing
R.~Senanayake, S.~O’Callaghan, and F.~Ramos, ``Predicting spatio-temporal propagation of seasonal influenza using variational gaussian process regression,'' \emph{Proceedings of the AAAI Conference on Artificial Intelligence}, vol.~30, no.~1, Mar. 2016. [Online]. Available: \url{https://ojs.aaai.org/index.php/AAAI/article/view/9899}
\BIBentrySTDinterwordspacing

\bibitem{8601344}
W.~Aftab, R.~Hostettler, A.~De~Freitas, M.~Arvaneh, and L.~Mihaylova, ``Spatio-temporal gaussian process models for extended and group object tracking with irregular shapes,'' \emph{IEEE Transactions on Vehicular Technology}, vol.~68, no.~3, pp. 2137--2151, 2019.

\bibitem{bonilla2007multi}
E.~V. Bonilla, K.~Chai, and C.~Williams, ``Multi-task gaussian process prediction,'' \emph{Advances in neural information processing systems}, vol.~20, 2007.

\bibitem{yu2005learning}
K.~Yu, V.~Tresp, and A.~Schwaighofer, ``Learning gaussian processes from multiple tasks,'' in \emph{Proceedings of the 22nd international conference on Machine learning}, 2005, pp. 1012--1019.

\bibitem{williams2008multi}
C.~Williams, S.~Klanke, S.~Vijayakumar, and K.~Chai, ``Multi-task gaussian process learning of robot inverse dynamics,'' \emph{Advances in neural information processing systems}, vol.~21, 2008.

\bibitem{osborne2008towards}
M.~A. Osborne, S.~J. Roberts, A.~Rogers, S.~D. Ramchurn, and N.~R. Jennings, ``Towards real-time information processing of sensor network data using computationally efficient multi-output gaussian processes,'' in \emph{2008 International Conference on information processing in sensor networks (ipsn 2008)}.\hskip 1em plus 0.5em minus 0.4em\relax IEEE, 2008, pp. 109--120.

\bibitem{durichen2014multi}
R.~D{\"u}richen, M.~A. Pimentel, L.~Clifton, A.~Schweikard, and D.~A. Clifton, ``Multi-task gaussian process models for biomedical applications,'' in \emph{IEEE-EMBS international Conference on Biomedical and health informatics (BHI)}.\hskip 1em plus 0.5em minus 0.4em\relax IEEE, 2014, pp. 492--495.

\bibitem{shen2023multi}
B.~Shen, R.~Gnanasambandam, R.~Wang, and Z.~J. Kong, ``Multi-task gaussian process upper confidence bound for hyperparameter tuning and its application for simulation studies of additive manufacturing,'' \emph{IISE Transactions}, vol.~55, no.~5, pp. 496--508, 2023.

\bibitem{9736401}
B.~Akbari and H.~Zhu, ``Tracking dependent extended targets using multi-output spatiotemporal gaussian processes,'' \emph{IEEE Transactions on Intelligent Transportation Systems}, vol.~23, no.~10, pp. 18\,301--18\,314, 2022.

\bibitem{NEURIPS2019_0118a063}
\BIBentryALTinterwordspacing
O.~Hamelijnck, T.~Damoulas, K.~Wang, and M.~Girolami, ``Multi-resolution multi-task gaussian processes,'' in \emph{Advances in Neural Information Processing Systems}, H.~Wallach, H.~Larochelle, A.~Beygelzimer, F.~d\textquotesingle Alch\'{e}-Buc, E.~Fox, and R.~Garnett, Eds., vol.~32.\hskip 1em plus 0.5em minus 0.4em\relax Curran Associates, Inc., 2019. [Online]. Available: \url{https://proceedings.neurips.cc/paper_files/paper/2019/file/0118a063b4aae95277f0bc1752c75abf-Paper.pdf}
\BIBentrySTDinterwordspacing

\bibitem{zhang2014power}
Y.~Zhang, G.~Luo, and F.~Pu, ``Power load forecasting based on multi-task gaussian process,'' \emph{IFAC Proceedings Volumes}, vol.~47, no.~3, pp. 3651--3656, 2014.

\bibitem{gilanifar2019multitask}
M.~Gilanifar, H.~Wang, L.~M.~K. Sriram, E.~E. Ozguven, and R.~Arghandeh, ``Multitask bayesian spatiotemporal gaussian processes for short-term load forecasting,'' \emph{IEEE Transactions on Industrial Electronics}, vol.~67, no.~6, pp. 5132--5143, 2019.

\bibitem{alber2019integrating}
M.~Alber, A.~Buganza~Tepole, W.~R. Cannon, S.~De, S.~Dura-Bernal, K.~Garikipati, G.~Karniadakis, W.~W. Lytton, P.~Perdikaris, L.~Petzold \emph{et~al.}, ``Integrating machine learning and multiscale modeling—perspectives, challenges, and opportunities in the biological, biomedical, and behavioral sciences,'' \emph{NPJ digital medicine}, vol.~2, no.~1, p. 115, 2019.

\bibitem{wang2019modeling}
D.~Wang, K.~Liu, and X.~Zhang, ``Modeling of a three-dimensional dynamic thermal field under grid-based sensor networks in grain storage,'' \emph{IISE Transactions}, vol.~51, no.~5, pp. 531--546, 2019.

\bibitem{raissi2019physics}
M.~Raissi, P.~Perdikaris, and G.~E. Karniadakis, ``Physics-informed neural networks: A deep learning framework for solving forward and inverse problems involving nonlinear partial differential equations,'' \emph{Journal of Computational physics}, vol. 378, pp. 686--707, 2019.

\bibitem{xie2022physics2}
J.~Xie and B.~Yao, ``Physics-constrained deep active learning for spatiotemporal modeling of cardiac electrodynamics,'' \emph{Computers in Biology and Medicine}, vol. 146, p. 105586, 2022.

\bibitem{cuomo2022scientific}
S.~Cuomo, V.~S. Di~Cola, F.~Giampaolo, G.~Rozza, M.~Raissi, and F.~Piccialli, ``Scientific machine learning through physics--informed neural networks: Where we are and what’s next,'' \emph{Journal of Scientific Computing}, vol.~92, no.~3, p.~88, 2022.

\bibitem{cai2021physics}
S.~Cai, Z.~Mao, Z.~Wang, M.~Yin, and G.~E. Karniadakis, ``Physics-informed neural networks (pinns) for fluid mechanics: A review,'' \emph{Acta Mechanica Sinica}, vol.~37, no.~12, pp. 1727--1738, 2021.

\bibitem{costabal2024delta}
F.~S. Costabal, S.~Pezzuto, and P.~Perdikaris, ``$\delta$-pinns: Physics-informed neural networks on complex geometries,'' \emph{Engineering Applications of Artificial Intelligence}, vol. 127, p. 107324, 2024.

\bibitem{krishnapriyan2021characterizing}
A.~Krishnapriyan, A.~Gholami, S.~Zhe, R.~Kirby, and M.~W. Mahoney, ``Characterizing possible failure modes in physics-informed neural networks,'' \emph{Advances in neural information processing systems}, vol.~34, pp. 26\,548--26\,560, 2021.

\bibitem{swiler2020survey}
L.~P. Swiler, M.~Gulian, A.~L. Frankel, C.~Safta, and J.~D. Jakeman, ``A survey of constrained gaussian process regression: Approaches and implementation challenges,'' \emph{Journal of Machine Learning for Modeling and Computing}, vol.~1, no.~2, 2020.

\bibitem{raissi2018hidden}
M.~Raissi and G.~E. Karniadakis, ``Hidden physics models: Machine learning of nonlinear partial differential equations,'' \emph{Journal of Computational Physics}, vol. 357, pp. 125--141, 2018.

\bibitem{raissi2018numerical}
M.~Raissi, P.~Perdikaris, and G.~E. Karniadakis, ``Numerical gaussian processes for time-dependent and nonlinear partial differential equations,'' \emph{SIAM Journal on Scientific Computing}, vol.~40, no.~1, pp. A172--A198, 2018.

\bibitem{pfortner2022physics}
M.~Pf{\"o}rtner, I.~Steinwart, P.~Hennig, and J.~Wenger, ``Physics-informed gaussian process regression generalizes linear pde solvers,'' \emph{arXiv preprint arXiv:2212.12474}, 2022.

\bibitem{dalton2024boundary}
D.~Dalton, A.~Lazarus, H.~Gao, and D.~Husmeier, ``Boundary constrained gaussian processes for robust physics-informed machine learning of linear partial differential equations,'' \emph{Journal of Machine Learning Research}, vol.~25, no. 272, pp. 1--61, 2024.

\bibitem{sorkine2005laplacian}
O.~Sorkine, ``Laplacian mesh processing,'' \emph{Eurographics (State of the Art Reports)}, vol.~4, no.~4, p.~1, 2005.

\bibitem{goldberger2000physiobank}
A.~L. Goldberger, L.~A. Amaral, L.~Glass, J.~M. Hausdorff, P.~C. Ivanov, R.~G. Mark, J.~E. Mietus, G.~B. Moody, C.-K. Peng, and H.~E. Stanley, ``Physiobank, physiotoolkit, and physionet: components of a new research resource for complex physiologic signals,'' \emph{circulation}, vol. 101, no.~23, pp. e215--e220, 2000.

\end{thebibliography}
}

\end{document}